\documentclass[review,authoryear,10pt]{elsarticle}

\usepackage{latexsym}
\usepackage{amssymb}
\usepackage[T1]{fontenc}
\usepackage{amsmath}
\usepackage{graphicx}
\usepackage{amsmath}
\usepackage{xcolor,float}
\usepackage{soul}
\usepackage[left=1in,top=1in,right=1in,bottom=1in,nohead]{geometry}
\usepackage{lineno}
\usepackage{color,soul}
\begin{document}
	
	\begin{frontmatter}
	\title{DeepCorn: A Semi-Supervised Deep Learning Method for High-Throughput Image-Based Corn Kernel Counting and Yield Estimation}
	%DeepCorn: A Deep Learning Method for Image-based Corn Kernel Counting and Yield Estimation
	\begin{abstract}
    The success of modern farming and plant breeding relies on accurate and efficient collection of data. For a commercial organization that manages large amounts of crops, collecting accurate and consistent data is a bottleneck. Due to limited time and labor, accurately phenotyping crops to record color, head count, height, weight, etc. is severely limited. However, this information, combined with other genetic and environmental factors, is vital for developing new superior crop species that help feed the world's growing population. Recent advances in machine learning, in particular deep learning, have shown promise in mitigating this bottleneck. In this paper, we propose a novel deep learning method for counting on-ear corn kernels in-field to aid in the gathering of real-time data and, ultimately, to improve decision making to maximize yield. We name this approach DeepCorn, and show that this framework is robust under various conditions. DeepCorn estimates the density of corn kernels in an image of corn ears and predicts the number of kernels based on the estimated density map. DeepCorn uses a truncated VGG-16 as a backbone for feature extraction and merges feature maps from multiple scales of the network to make it robust against image scale variations.  We also adopt a semi-supervised learning approach to further improve the performance of our proposed method. Our proposed method achieves the MAE and RMSE of 41.36 and 60.27 in the corn kernel counting task, respectively. Our experimental results demonstrate the superiority and effectiveness of our proposed method compared to other state-of-the-art methods.
	\end{abstract}
	
	\begin{keyword}
	 Corn kernel counting, Convolutional neural network, Semi-supervised learning, Deep learning, Yield estimation\\
	 \vspace{1cm}
	  \textbf{This manuscript is published in the Knowledge-Based Systems journal, please visit https://doi.org/10.1016/j.knosys.2021.106874 for getting the publisher version of the manuscript.}
	 
	 \textbf{This manuscript version is made available under the CC-BY-NC-ND 4.0 license http://creativecommons.org/licenses/by-nc-nd/4.0/.}
	\end{keyword}

	\author[ISU]{S.~Khaki\corref{cor1}}
	\ead{skhaki@iastate.edu}	
	
	\author[SYN]{H.~Pham}
	%\ead{hieu.pham@syngenta.com}
		
	\author[SYN]{Y.~Han}
	%\ead{ye.han@syngenta.com}
	
	\author[SYN]{A.~Kuhl}
	%\ead{andy.kuhl@syngenta.com}
	
	\author[SYN]{W.~Kent}
	%\ead{wade.kent@syngenta.com}
	
	\author[ISU]{L.~Wang}
	%\ead{lzwang@iastate.edu}
	
	%\cortext[cor1]{Corresponding author: S. Khaki}
	
	\address[ISU]{Department of Industrial and Manufacturing Systems Engineering, Iowa State University, Ames, Iowa USA}
	
	\address[SYN]{Syngenta Seeds, Slater, Iowa USA}	
	
	\end{frontmatter}

\section{Introduction}

High throughput phenotyping (HTP), the  is a limiting factor facing modern agriculture. Due to labor intensive tasks, phenotyping crops to identify color, stand count, leaf count, plant height, etc. is severely limited. This ``phenotying bottleneck'' restricts our capability to examine how phenotypes interact with the plant's genetics factors as well as environmental factors \citep{furbank2011phenomics}. For a farmer who manages upwards of 10,000 acres, it is infeasible to be able to inspect each crop individually to identify its characteristics. In an ideal scenario, HTP results in the collection, annotation, and labeling of massive amounts of data for analysis that is vital for advancing plant breeding. Unlike other domains, live, in-field data can only be collected at a specific period during a plant's growth cycle. If this time is missed, then a farmer or breeder must wait until the next growing period to collect more data which, in some cases, may be one year later. To mitigate this issue, agronomists have turned to image-based capturing techniques (such as phones and drones) as a means of data collection and storage. Through these images, farmers no longer are bound by a plant's growth cycle and can thus phenotype a crop at a later date. However, with the influx of a massive amount of image-based data, farmers and breeders are now faced with a similar but new challenge - analyzing massive amounts of images quickly. To effectively perform image-based HTP, tools must be made available to farmers and breeders to make real-time decisions to manage their crops against pests, disease, drought, etc. to maximize their growth and, ultimately, yield. 

Modern agriculture has evolved to encompass drone, satellite, and cellphone imagery as a method for data storage and collection. The purpose of these technologies is to capture still-images so that identification of plant characteristics can be analyzed at a later date. Although this helps in mitigating the data collection phase of HTP, this now creates a new problem in being able to accurately and efficiently analyze the captured image to obtain the desired information. Recent years have seen advancements at the intersection of planting phenotyping and traditional machine learning approaches to identify  stress, coloring, and head count in soybeans, wheat, and sorghum \citep{singh2016machine, naik2017real, yuan2018wheat, guo2018aerial}. These studies showcase and emphasize the impact the HTP can have on advancing our knowledge of plants and their interaction with the environment as well as how to make effective real-time decisions to protect a crop during its growing season. Although traditional machine learning approaches have seen success in agriculture, the current state of the art in HTP is with the application of deep learning.

Deep learning techniques in agriculture is used for image classification, object detection and counting. Common deep learning frameworks such as AlexNet \citep{krizhevsky2017imagenet}, LeNet \citep{lecun1998gradient}, and VGG-16 \citep{simonyan2014deep} have been applied to classify diseases in tomatoes, cherries, apples,  peppers, and olives \citep{mohanty2016using, cruz2017x,wang2017automatic}. Various papers have also utilized traditional object detection architectures such as ResNet50 to count leaves and sorghum heads \citep{giuffrida2018pheno, mosley2020image}. Outside of existing frameworks, \cite{lu2017tasselnet} created a novel approach to identify corn tassels by combining convolutional neural networks (CNN) and local counts regression into what they refer to as TasselNet. In addition to various applications, numerous annotated datasets across different crops have been made publicly available to researchers for classification and detection problems \citep{zheng2019cropdeep,sudars2020dataset,haug2014crop}. Indeed, these works show how rapidly the combination of deep learning and plant breeding is growing and provide hope in mitigating the bottleneck facing the analysis of crop images.  For a comprehensive overview of image-based HTP we refer the reader to a paper by \cite{jiang2020convolutional}.

Although there is vast literature covering various crops and object detection approaches, there are few studies that perform HTP on commercial corn (\textit{Zea mays} L.). Previous studies proposed deep learning based methods to accurately predict corn yield based on factors such as genotype, weather, soil, and satellite imagery, but non of these studies are considered as the HTP on commercial corn \citep{khaki2019classification,russello2018convolutional,khaki2019cnn,khaki2019crop,khaki2020predicting}. Due to the number of food and industrial products that dependent on it, corn is widely regarded as one of the world's most vital crops \citep{berardi2019flooding}. Not only is corn used to create products such as starch, flour, and ethanol, it is also the primary feed for livestock (pigs, cows, cattle, etc.) due to being rich in nutrients and proteins \citep{nazli2018potential}. In 2019, corn was the United States' (U.S.) largest grown crop accounting for more than 90 million acres of land and adding more than \$140 billion to the U.S. economy \citep{usda2019}. By 2050, the world's population is expected to reach 9.1 billion \citep{stephenson2010population}. With the world's population increasing and the amount of arable land non-increasing, changes must occur to maximize corn yield while maintaining the same (or fewer) input parameters. 

A practical approach to increasing corn yield is to create a way for agronomists and farmers to have a real-time, precise mechanism to estimate yield during the growing season. Such a mechanism would enable farmers to make real-time grain management decisions to maximize yield and minimize potential losses of profitability. By having an estimate on yield, farmers are able to decide optimal management practices (when to apply fungicide, nitrogen, fertilizer, etc.) to aid the yield potential of corn. Currently, the process of estimating in-season yield relies on manual counting of corn kernels and using certain formulas to get an approximated yield per acre. However, a healthy ear of corn contains 650 - 800 kernels. Therefore individually counting each kernel on an ear is a labor-intensive, time-consuming, and error prone task. Moreover, to provide an accurate representation of a field's true yield, a farmer would need to count kernels on multiple ears, further adding to the labor and time requirements. To aid in solving this HTP bottleneck, \cite{khaki2020convolutional} developed a sliding window convolutional neural network (CNN) approach to count corn kernels from an image of a corn ear. However, their approach did not estimate yield. Moreover, their proposed approach required a fixed distance between ears and camera and was not invariant to the ear orientation and the kernel color. Their sliding window approach also increased the inference time. Because of these limitations, their approach is not suitable for large scale deployment.

Given the need to effectively and efficiently count corn kernels to estimate yield, we present a novel approach to solve this by utilizing a new deep learning architecture, which we name DeepCorn. The problem of counting kernels is similar to counting dense groups of people in crowds due to the compactness and the varying angles and sizes of kernels. As such, we will be comparing our method to commonly used crowd counting methods in the literature, which are applicable to other counting studies. Specifically, the novelties of our approach include:

\begin{enumerate}
    \item A robust on-ear kernel counting that is invariant to image scale variation, ear shape, size, orientation, lighting conditions, and color
    \item A new deep learning architecture that is superior to commonly used crowd counting models
    \item The utilization of a semi-supervised learning approach to further improve the performance of the proposed method. To the best of our knowledge, our paper is the first to generate pseudo-density maps of unlabeled images to improve the counting accuracy in the literature of crowd counting \citep{gao2020cnn}.
    \item Proposing an approach to effectively and efficiently estimate corn yield based on the output of our proposed method
    \item A kernel counting dataset to benchmark our proposed method
    
\end{enumerate}

%This paragraph talk about similar deep learning problems like crowd counting--

To achieve this goal, in Section \ref{method} we provide an overview of our deep learning architecture. Section \ref{experiments} provides the details about our experimental setup, dataset and annotations, evaluation metrics and benchmark models. Analysis is performed in Section 4 to study the robustness of our framework and a procedure for estimating yield. Lastly, Section 5 concludes with our key results and findings.

Our proposed method is motivated by the density estimation-based crowd counting methods since they both have the goal of counting a large number of densely packed objects in an image . Crowd counting methods aim to estimate the number of people in an image, which is challenging due to large image scale variations, background clutters, and occlusions \citep{gao2020cnn, cao2018scale}. Recently, CNN-based density map estimation methods have shown great promise in crowd counting task \citep{liu2018crowd,liu2019context,ma2019bayesian,liu2019crowd,liu2020efficient}. For example, \cite{liu2020efficient} proposed a novel structured knowledge transfer framework which uses the structured knowledge of a trained teacher network for generation of a lightweight student network for crowd counting. \cite{liu2019crowd} designed a deep structured scale integration network to counter the image scale variations using structured
feature representation learning and hierarchically structured loss function optimization. Recent state-of-the-art crowd counting methods also have proposed context-aware and Bayesian loss for crowd counting which proved to be highly successful \citep{ma2019bayesian, liu2019context}. 

\section{Methodology}\label{method}

The goal of this study is to count corn kernels in an image of corn ears taken in uncontrolled lighting conditions, and, ultimately, use the number of counted kernels to estimate yield. In this paper, we propose a deep learning based method, DeepCorn, to count all the corn kernels given a single 180-degree image of a corn ear. With this single angle image, we aim to estimate the number of kernels on the entire corn ear. Although, as shown later in our study, having multiple images from different sides of the ear is beneficial, there are time and automation considerations. From the perspective of the candidate farmer or plant breeder, multiple images may be taken at the expense of a few more seconds of image capturing. However, from an automation perspective where ears are on a conveyor belt before kernels are shelled off the ear, a single image is all that is manageable. Given these considerations and our goal of creating a robust, generalizable model that is applicable in different processes, we focus our attention on the case where only a single image at a single angle is given for an ear of corn.

\begin{figure}[H]
    \centering
    \includegraphics[scale=0.05]{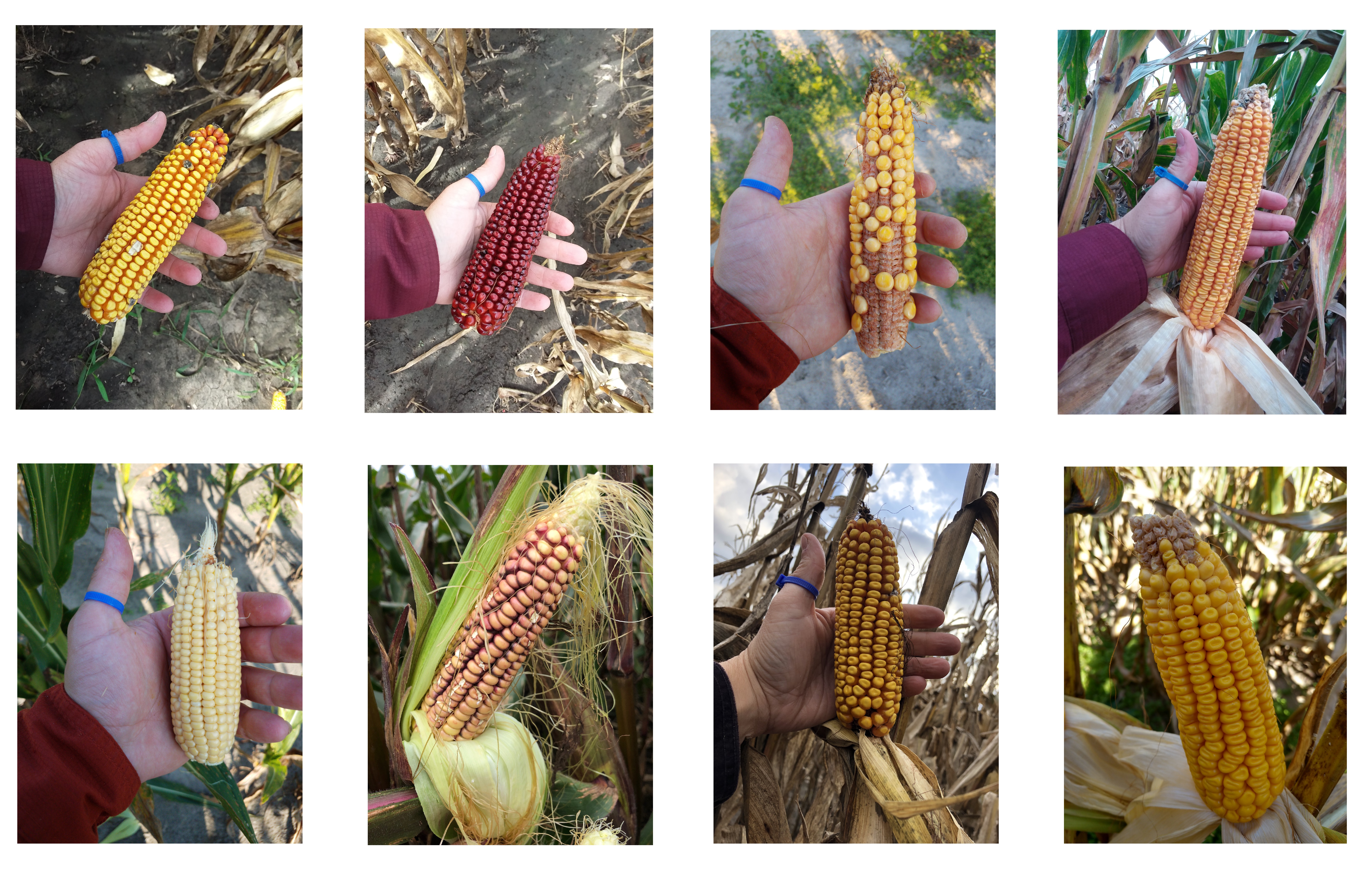}
    \caption{Eight genetically different corn ears. The images indicate the scale variations and the genetic difference among ears.}
    \label{fig:8_ears}
\end{figure}

Image-based corn kernel counting is challenging due to multiple factors: (1) large variation in kernel shapes and colors (yellow, red, white, etc), (2) very small distance between kernels, and (3) different orientations, lighting conditions, and scale variations of the ears. Figure \ref{fig:8_ears} displays eight genetically different corn ears which illustrates these factors.  Our proposed model is inspired by the crowd counting models because they both aim to count a large number of densely packed objects in an image \citep{gao2020cnn}. 

\subsection{Network Architecture}

\begin{figure}[H]
    \centering
    \includegraphics[scale=0.22]{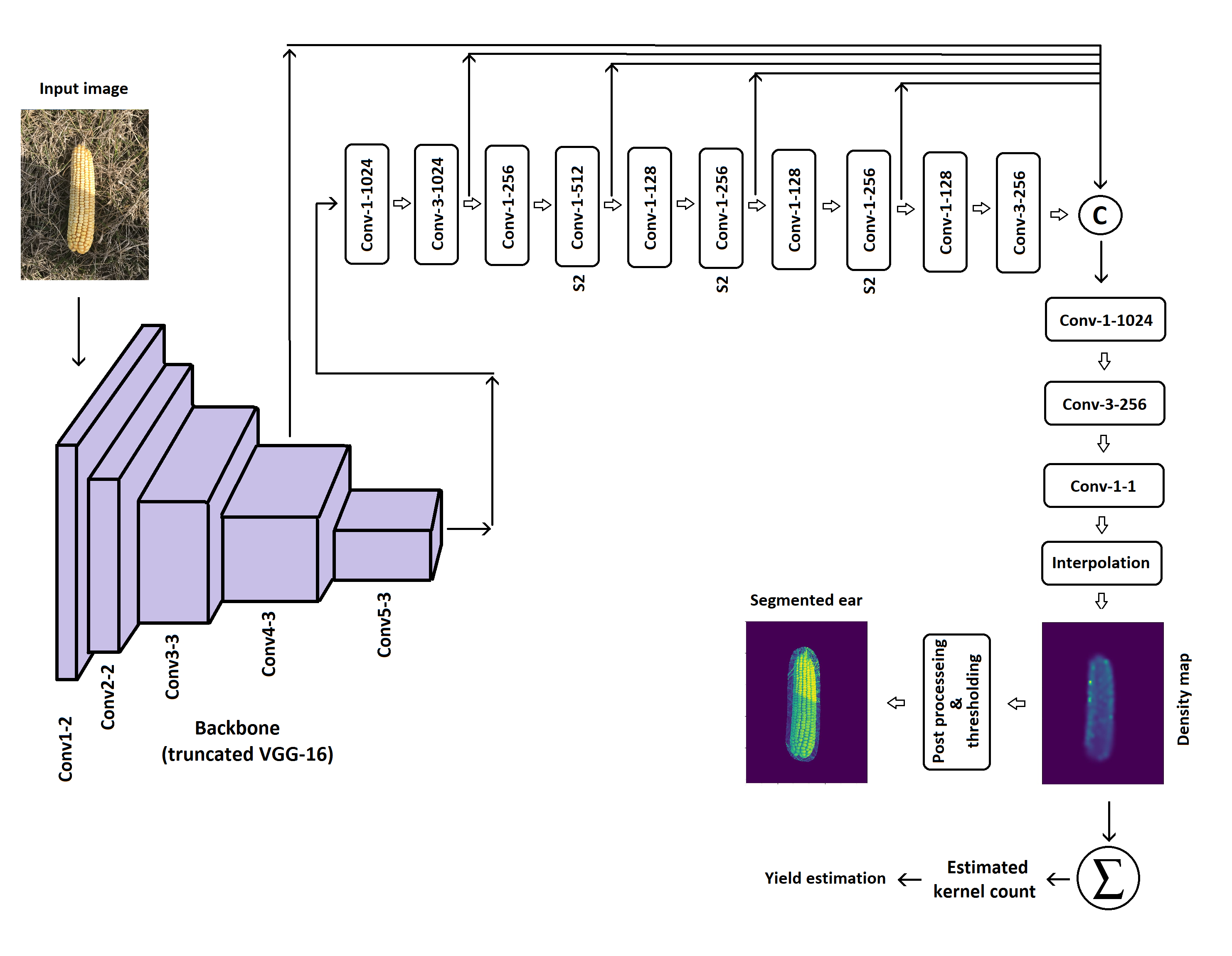}
    \caption{Outline of the DeepCorn architecture. The parameters of the convolutional layers are denoted as ``Conv-(kernel size)-(number of filters)''. The amount of stride for all layers is 1 except for the layers with ``S2'' notation for which we use stride of 2. The padding type is `same' for all convolutional layers except the last layer before the concatenation, which has `valid' padding. \textcircled{\raisebox{-0.8pt}{c}} and $\Sigma$ denote matrix concatenation and summation over density map, respectively.  }
    \label{fig:diagram}
\end{figure}

Corn ear images include various sizes of kernel pixels due to the image scale variations and having genetically different corn kernel shapes and colors. As such, the proposed method should be able to counter scale variations and learn both highly semantic levels (ears, background, etc.) and low-level patterns (kernel edges, colors, etc.). Figure \ref{fig:diagram} outlines the architecture of the proposed method. The proposed network is used to estimate the image density map whose integral over any region in the density map gives the count of kernels within that region. Various state-of-the-art approaches utilize a density map construction approach to count dense objects in crowds and have shown to be highly successful \citep{liu2018crowd,liu2019context,ma2019bayesian, liu2019crowd,liu2020efficient}. We use density estimation-based approach rather than detection-based or regression-based approaches for the following reasons. Detection-based approaches usually apply an object detection method like sliding window \citep{khaki2020convolutional} or more advanced methods such as YOLO \citep{redmon2016you}, SSD \citep{liu2016ssd}, and fast R-CNN \citep{girshick2015fast} to first detect objects and subsequently count them. However, their performance is unsatisfactory in dense object counting \citep{gao2020cnn} and they also require huge amount of annotated images, which may not be publicly available for corn kernel counting. Regression-based approaches \citep{wang2015deep,chan2008privacy,idrees2013multi,chan2009bayesian} are trained to map directly an image patch to the count. Despite being successful in counting problems such as occlusion and background clutter, these methods have a tendency to ignore spatial information \citep{gao2020cnn}.

Our proposed network uses VGG-16 \citep{simonyan2014deep} as a backbone
for feature extraction. Originally proposed for image classification, the VGG-16 network stacks convolutional layers with a fixed kernel size of $3\times3$, which usually generalizes well to
other vision tasks including object counting and detection \citep{shi2018multiscale,boominathan2016crowdnet,gao2020counting,sang2019improved,valloli2019w,liu2016ssd,kumar2019mtcnet}. We exclude the last max-pooling layer and all fully connected from the VGG network. Even though, the VGG-16 network was originally designed to process input image with size of $224\times224$, we use input image with size of $300\times300$ because the proposed network can potentially learn more fine-grained patterns with higher resolution input images \citep{tan2019efficientnet}.

To make the proposed model robust against scale variations and perspective change in images, we merge feature maps from multiple scales of the network. As such, the model can easily adapt to the scale and perspective changes. Similar scale-adaptive CNN approaches have also been used in other counting and detection studies \citep{zhang2018crowd,sang2019improved,liu2016ssd}. Compared to other studies \citep{zeng2017multi,cao2018scale,boominathan2016crowdnet,sam2017switching,zhang2016single,deb2018aggregated} that used multi-column architecture with different filter sizes to deal with scale variations, our proposed network has fewer parameters due to parameter sharing, which can accelerate the training process. Moreover, these skip connections can prevent the vanishing gradient problem and further accelerate the training process as recommended in \citep{he2015deep}. To concatenate the feature maps from multiple scales, we increase the spatial dimensions of the feature maps to have the same size as the largest feature map (first feature map) using zero padding, which is the concatenation approach recommended in \citep{he2015deep}. After concatenation, feature maps are passed to two convolutional layers. We use $1\times1$ convolutional layers throughout our network to increase or decrease its depth without a significant performance penalty \citep{szegedy2015going}.

Due to having four max-pooling layers in our network, the special resolution of the last convolutional layer of the network is of $1/8$ of the input image. Finally, we up-sample the output of the last convolutional layer to the size of the input image using bi-linear interpolation to estimate the density map. The total count of kernels in the image can be obtained by a summation over the estimated density map. Finally, we put a threshold on the estimated density map to zero-out regions where the value of density map is very small. Then, we combine the thresholded density map with the input image which results in segmented corn ears.

%n order to have a better visualization, 
\subsection{Kernel Counting Using Proposed Model}

The ground truth density maps are generated in our study in a way that the summation over the density map is the same as the total number of kernels in the image. Such property enables the proposed model to count the number of kernels in an image as an auxiliary task while learning how much each region of the image contributes to the total count. As a result, the proposed method is trained end-to-end to predict the density maps where the number of kernel can be easily estimated by summation over the predicted density maps at the inference time.

\subsection{Network Loss}

We employ the Euclidean loss as shown in Equation \ref{eq:eq_loss} to train the network. The Euclidean loss is a popular loss in crowd counting literature due to enhancing the quality of the estimated density map \citep{gao2020counting,boominathan2016crowdnet,shi2018multiscale,wang2019removing};

\begin{eqnarray}
L(\theta)=\frac{1}{N}\sum_{i=1}^{N}\|F(X_i,\Theta)-D_i\|_{2}^{2}\label{eq:eq_loss}
\end{eqnarray}

\noindent where, $F(X_i,\Theta)$, $\theta$, $X_i$, $D_i$, and $N$ denote the predicted density map of the $i$th input image, the parameters of the network, the $i$th input image, the $i$th ground truth density map, and the number of images, respectively. Euclidean loss measures the distance between the estimated density map and the ground truth. 

The Euclidean loss computes the difference between the predicted and ground truth density maps at the pixel level and then sums over to compute the loss for each image.

\section{Experiment and Results}\label{experiments}

In this section, we present the dataset used for our experiments, the evaluation metrics, the training hyper-parameters, and final results. We conducted all experiments in Tensorflow \citep{abadi2016tensorflow} on a NVIDIA Tesla V100 GPU.

\subsection{Dataset}

\subsubsection{Ground Truth Density Maps}

We follow the procedure in \citep{boominathan2016crowdnet} to generate ground truth density maps. If we assume the center of one corn kernel is located at pixel $x_i$, then the kernel can be represented by a delta function $\delta(x-x_i)$. As such, the ground truth for an image with $M$ kernel center annotations can represented as follows:

\begin{eqnarray}
H(x)=\sum_{i=1}^{M}\delta(x-x_i)\label{eq:eq_1}
\end{eqnarray}

\noindent Then, $H(x)$ is convoluted with a Guassian kernel to generate the density map $D(x)$: \\
\begin{eqnarray}
D(x)=\sum_{i=1}^{M}\delta(x-x_i) \ast G_\sigma(x)\label{eq:eq_2}
\end{eqnarray}

\noindent where $\sigma$ denotes the standard deviation. The parameter $\sigma$ is determined based on the average distance of $k$-nearest neighboring kernel annotations. The summation over the density map is the same as the total number of kernels in the image. Using such a density map can be extremely beneficial as it helps the CNN learn how much each region contributes to the total count.

\subsubsection{Input Images and Data Augmentation}\label{sec:data}

 We perform the following procedure to prepare the training data for our proposed CNN model. We use 109 corn ear images with a fixed size of $1024\times768$ as our training data. Table \eqref{tab:data_stat} shows the statistics of the dataset. The statistics reported in Table \ref{tab:data_stat} are for one side of corn ears which was captured in images.
 
 \begin{table}[H]
     \centering
     \begin{tabular}{|c|c|c|c|c|c|}
     \hline
       Number of Images&  Resolution  & Min & Max& Avg & Total  \\
         \hline
          109& $1024\times768$  & 16 & 1,116 & 182.02 & 19,848\\ 
         \hline
     \end{tabular}
     \caption{The statistics of dataset used in this study. Min, Max, Avg, and Total denote the minimum, maximum, average, and total kernel numbers, respectively. }
     \label{tab:data_stat}
 \end{table}

To better train the proposed CNN model, we augment the dataset in the following way. First, we construct the multi-scale pyramidal representation
of each training image as in \cite{boominathan2016crowdnet} by considering scales of 0.4 to 1.6, incremented in steps of 0.1, times the original image resolution. Then, we randomly crop 80 patches of size $300\times300$ pixels from each scale of image pyramid. Such augmentation makes the proposed model robust against scale variations. In order to make the proposed model robust against orientation of ears and lighting condition, we perform extensive augmentations such as random rotation, flip, adding Gaussian noise, and modifying brightness and contrast on 40\% of the randomly selected cropped image patches.

\subsection{Evaluation Metrics}

To evaluate the performance of our proposed model, we use the mean absolute error (MAE), root mean squared error (RMSE) metrics, and mean absolute percentage error (MAPE), which are defined as follows:\\

\begin{eqnarray}
MAE=\frac{1}{N}\sum_{i=1}^{N}|C_{i}^{GT}-C_{i}^{pred}|\label{eq:MAE_eq}
\end{eqnarray}

\begin{eqnarray}
RMSE=\sqrt{\frac{1}{N}\sum_{i=1}^{N}|C_{i}^{GT}-C_{i}^{pred}|^2}\label{eq:RMSE_eq}
\end{eqnarray}

\begin{eqnarray}
MAPE=\frac{1}{N}\sum_{i=1}^{N}|\frac{C_{i}^{GT}-C_{i}^{pred}}{C_{i}^{GT}}|\times 100\label{eq:MAPE_eq}
\end{eqnarray}

\noindent where, $N$, $C_{i}^{pred}$, and $C_{i}^{GT}$ denote the number of test images, the predicted counting for $i$th image, and the ground truth counting for $i$th image, respectively.

\subsection{Semi-supervised Learning}

Semi-supervised learning utilizes both labeled and unlabeled data during the training step. The amount of labeled data is often limited and scarce in many learning tasks. As a result, semi-supervised learning methods can leverage unlabeled data to improve the learning accuracy. For example, \cite{caron2018deep} proposed a clustering method called DeepCluster which jointly learns the parameters of the convolutional neural network and the cluster assignment. Their proposed method uses the cluster assignment as pseudo-labels to learn the weights of the network. \cite{wu2019progressive} developed a progressive framework for person re-identification based on only one labeled example. Their proposed framework generates pseudo-labeled data and uses them along with original labeled data to train the CNN network. In another study, \cite{xie2020self} proposed a self-training method which generates pseudo labeled images and uses them with labeled images during the training to further improves the learning accuracy.

To improve the performance of our proposed method, we adopt a semi-supervised learning approach to generate pseudo-density maps and use them to train our proposed method. We use the noisy student training algorithm proposed by \cite{xie2020self} which is as follows:

\begin{enumerate}
    \item We train the proposed model, called teacher, on the labeled dataset $\{(X_i,D_i), \,  i=1,...,n\}$, where $X_i$, $D_i$, and $n$ are the $i$th labeled image, the $i$th ground truth density map, and the number of labeled images, respectively.
    \item We use the trained teacher model denoted as $F$ to generate pseudo-density maps, denoted as $\tilde{D}$, for the unlabeled dataset $\{\tilde{X_j}, j=1,...,m\}$, where $\tilde{X_j}$ and $m$ are the $j$th unlabeled image and the number of unlabeled images, respectively. 
    
    $\tilde{D_j}=F(\tilde{X_j}), \,  j=1,...,m$
    
    \item Finally, we retrain the proposed model with noise, called student, using both labeled and pseudo-labeled images.
    
\end{enumerate}

In all, we use this semi-supervised learning approach to do a two-level training where the teacher learns on actual labelled images and then is employed to generate pseudo-labeled images. In the end,  the student model learns on both labeled and pseudo-labeled images with the intent that the student model is better than the teacher.

\subsection{Training Details}

\begin{figure}[H]
    \centering
    \includegraphics[scale=0.4]{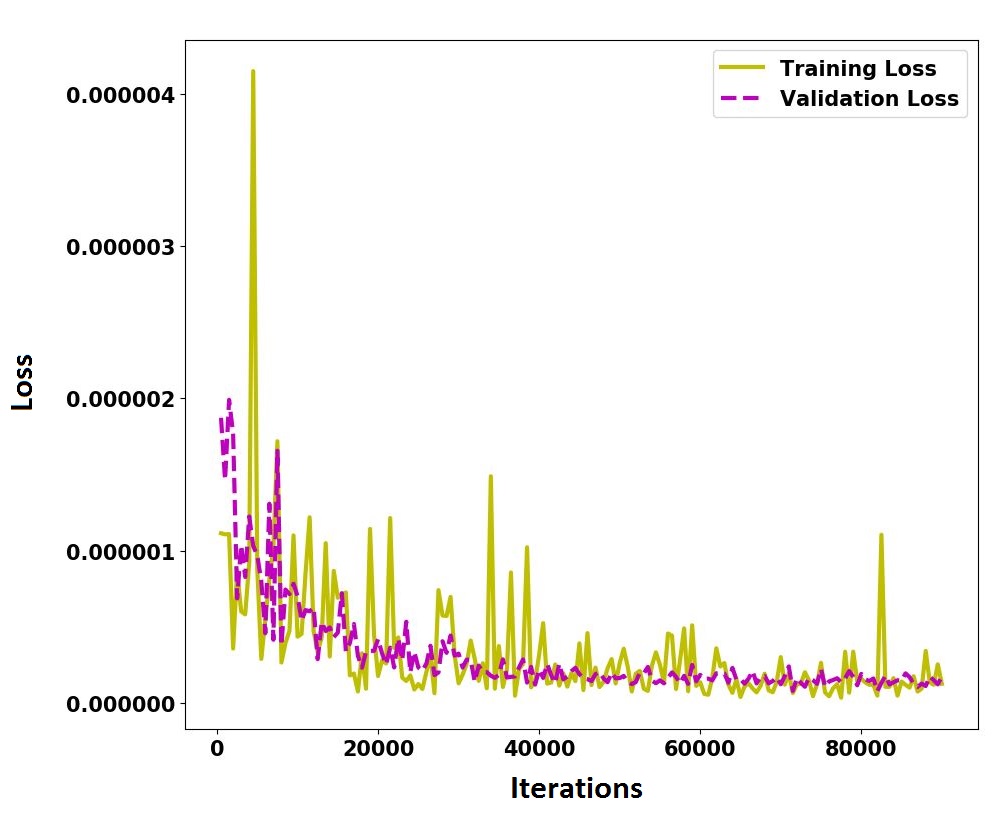}
    \caption{Plot of training and validation losses of the DeepCorn model during training process.}
    \label{fig:loss_plot}
\end{figure}

We train our proposed model end-to-end using the following training parameters. We apply the data augmentation described in Section \ref{sec:data} on our dataset, which resulted in 154,169 image patches. We randomly take 90\% of image patches as training data and use the rest as validation data to monitor the training process. We initialize the weights of network with the Xavier initialization \citep{glorot2010understanding}. Stochastic gradient descent (SGD) is used with a mini-batch size of 16 to optimize the loss function defined in Equation \eqref{eq:eq_loss} using Adam optimizer \citep{kingma2014adam}. We set the learning rate to be 3e-4 which was gradually decayed to 2.5e-5 during training. The model was trained for 90,000 iterations. Figure \ref{fig:loss_plot} shows the plot of training and validation losses during the training process.

For the semi-supervised Learning part, we first used our trained model as a teacher and generated pseudo density maps for 1000 unlabeled images of corn ears. Then, we added input noises including random rotation, flip and color augmentations to make a noisy dataset of 30,000 pseudo labeled images. We trained the student model with a mini-batch size of 16 which includes 2 samples from pseudo labeled images and 14 samples from original labeled dataset. We trained the student model for 200,000 iterations.  

\subsection{Design of Experiments}

To evaluate the counting performance of our proposed model, we compare our proposed model with ten state-of-the-art models originally proposed for crowd counting, but applicable to other counting problems with dense objects, which are as follows:\\
\noindent \textbf{DensityCNN:} proposed by \cite{jiang2020density}, this model uses a density-aware CNN which utilizes high-level semantic information to provide guidance and constraint when generating density maps. Their proposed method adopts a multi-task group-based CNN structure to jointly learn density-level classification and density map estimation.

\noindent \textbf{SCAR:} proposed by \cite{gao2019scar}, this model utilizes a CNN with spatial/channel-wise attention modules to estimate the density maps. Spatial-wise attention module of their proposed model encodes the pixel-wise context of the entire image to more accurately predict density maps at the pixel level while the channel-wise attention module extracts more discriminative features among different channels.  

\noindent \textbf{ACSPNet:} proposed by \cite{ma2019atrous}, this model employs an atrous convolutions spatial pyramid network for density estimation. Their proposed CNN model uses atrous convolutions to exaggerate the receptive field and maintain the resolution of extracted features. Their proposed model also uses atrous spatial pyramid pooling to merge features at different scales to counter image scale variations.

\noindent \textbf{ASD:} proposed by \cite{wu2019adaptive}, this model proposes an adaptive scenario discovery framework for density estimation. Their proposed model has two parallel
sub-networks that are trained with different sizes of the receptive
field to represent different scales and object densities. Their proposed model also adaptively assigns weights to the output of two sub-networks' responses by discovering and modeling the dynamic scenarios implicitly.

\noindent \textbf{SPN:} proposed by \cite{chen2019scale}, this model uses a CNN with scale pyramid network structure which uses a shared single deep column structure to extract multi-scale information in high layers by a scale pyramid module. Their proposed model adopts different rates of dilated convolutions in parallel in scale pyramid module to make their model robust against image scale variations.

\noindent \textbf{SaCNN:} proposed by \cite{zhang2018crowd}, this model uses a scale-adaptive CNN architecture with VGG backbone to estimate the density map. The SaCNN model merges feature maps from two different scales of the network to make the proposed network robust against the scale variation.

\noindent \textbf{CSRNet:} proposed by \cite{li2018csrnet}, this model includes VGG backbone as the front-end for 2D feature extraction and and a dilated CNN for the back-end. CSRNet adopts dilated convolutional layers to enlarge receptive fields to replace pooling operations. In the end, the output of the network is upsampled to estimate the density map.

\noindent \textbf{MSCNN:} proposed by \cite{zeng2017multi}, this model uses inception-like modules to extract scale-relevant features in their CNN network architecture and estimate the density map. The inception-like module is composed of multiple filters with different kernel size to make the model robust against scale variation. 

\noindent \textbf{CrowdNet:} proposed by \cite{boominathan2016crowdnet}, this model combines a deep CNN and a shallow CNN to predict the density map. The CrowdNet uses VGG as a deep network to extract high-level semantics features and a three-layer shallow network to extract low-level features. This network design makes the model robust against the scale variation.

\noindent \textbf{DeepCrowd:} proposed by \cite{wang2015deep}, this model is a regression based approach which directly maps the input image to the count. The DeepCrowd uses a custom CNN architecture which includes five convolutional layers and two fully connected layers at the end.

\subsection{Final Results}

Having trained our proposed model, we can now evaluate the performance of our proposed model to count corn kernels. To estimate the number of kernels on a whole corn ear using a 180-degree image, we count the number of kernels on one side of the ear, and then double it to estimate the total number of corn kernels on a corn ear, because of physiological factors, farmers and agronomists assume that corn ears are symmetric \citep{bennetzen2008handbook}. However, we empirically found that the multiplier coefficient of 2.10 works best for approximating the kernels on the both side of an ear from a 180 degree image.

To evaluate the performance of our proposed method, we manually counted the entire number of kernels on 291 different corn ears. Specifically, for our evaluation, the ground truth for our 291 corn ears and the remainder of this paper is the number of kernels on the entire ear. We use our proposed model along with the models described in the design of experiment section to predict the number of kernels on these corn ears given a single image from a single side of the ear. The competing models are trained on our corn dataset described in \ref{sec:data} and we report the results on the testing set. The test data include many difficult test images such as non-symmetric corn ears and uncontrolled lighting conditions. Table \ref{tab:result1} compares the performances of the competing methods with respect to the evaluation metrics. We report the performance of two versions of our proposed model, namely teacher and student models. The teacher model is trained on the original labeled data while the student model is trained on the both labeled and pseudo labeled data.

\begin{table}[H]
    \centering
    \begin{tabular}{|c|c|c|c|c|}
    \hline
        Method &  MAE & RMSE & MAPE &\begin{tabular}{c}
             Number of \\
             Parameters (M)\\
        \end{tabular}\\
         \hline
         DensityCNN \citep{jiang2020density} & 50.76&67.98 &52.83& 18.26\\
         \hline
         SCAR \citep{gao2019scar}&47.95&68.02 & 35.40&16.27\\
         \hline
         ACSPNet \citep{ma2019atrous} & 55.13&75.04&43.93&1.81\\
         
         \hline
         
         ASD \citep{wu2019adaptive}& 50.09&67.84&38.45&50.86\\
         \hline
         SPN \citep{chen2019scale} & 50.21 &64.21&35.82&32.41\\
   
         \hline
        SaCNN \citep{zhang2018crowd}  & 49.14 &68.30&50.34&25.07 \\
       
        \hline
        CSRNet \citep{li2018csrnet}  & 56.52& 74.21&54.65&16.26\\
        \hline
        
        MSCNN \citep{zeng2017multi}  & 56.07& 74.94&60.61&3.08 \\
        \hline
        CrowdNet \citep{boominathan2016crowdnet}  &90.50 &120.21&44.20&14.75 \\
        \hline
        DeepCrowd \citep{wang2015deep} & 93.11 &112.56 &87.73&71.92 \\
        \hline
        Our proposed (teacher)  & 44.91 & 65.92&37.87&26.61 \\
        \hline
         Our proposed (student)  & \textbf{41.36} & \textbf{60.27}&\textbf{35.38}&26.61 \\
        \hline
    \end{tabular}
    \caption{The comparison of the competing methods on kernel counting performance on the 291 corn ears. }\label{tab:result1}
\end{table}

As shown in Table \ref{tab:result1}, the proposed method outperformed the other methods to varying extents. The student model outperformed the teacher model due to using semi-supervised learning which helps the proposed model generalize better to test dataset. The SaCNN performed better than other methods except our proposed method and SCAR due to using scale-adaptive architecture and merging feature maps from two different scales. Such an architecture makes the SaCNN more robust against scale variation. CSRNet, ACSPNet, and MSCNN had a comparable performance and outperformed the CrowdNet and DeepCrowd methods. The SCAR method outperformed other methods except our proposed method with respect to MAE due to using spatial/channel wise attention modules in their network which help extract more discriminative features. The DensityCNN, ASD, SaCNN, and SPN had a similar performance, however, SaCNN had a lower MAE compared to these methods.

The proposed method performed considerably better than other methods due its deep architecture and merging multiple feature maps from different depths of network, which make the model significantly robust against scale variation. The DeepCrowd method as a regression-based method had similar performance as the CrowdNet method. At the test time, we have to crop a test image into some non-overlapping patches and feed them as input to the all methods except for our proposed method and CrowdNet. Then, we assemble the corresponding estimated density maps to obtain the final total count of the image. Otherwise, the performances of these method are not satisfactory when the whole image is fed to these method. But, our proposed method and CrowdNet take the whole image as input, and output the corresponding density map in a single forward path. Therefore, the inference time of the DeepCorn and CrowdNet are significantly smaller than other methods. The highest inference time belongs to the ACSPNet method because it does not use any downsampling in its network architecture and performs convolution operations on high resolution image throughout the network. The inference time of the proposed method is 0.65 second on an Intel i5-6300U CPU 2.4 GHz. Therefore, our proposed method achieves highest prediction accuracy while having considerably small inference time and moderate amount of parameters.

\begin{figure}[H]
    \centering
    \includegraphics[scale=0.20]{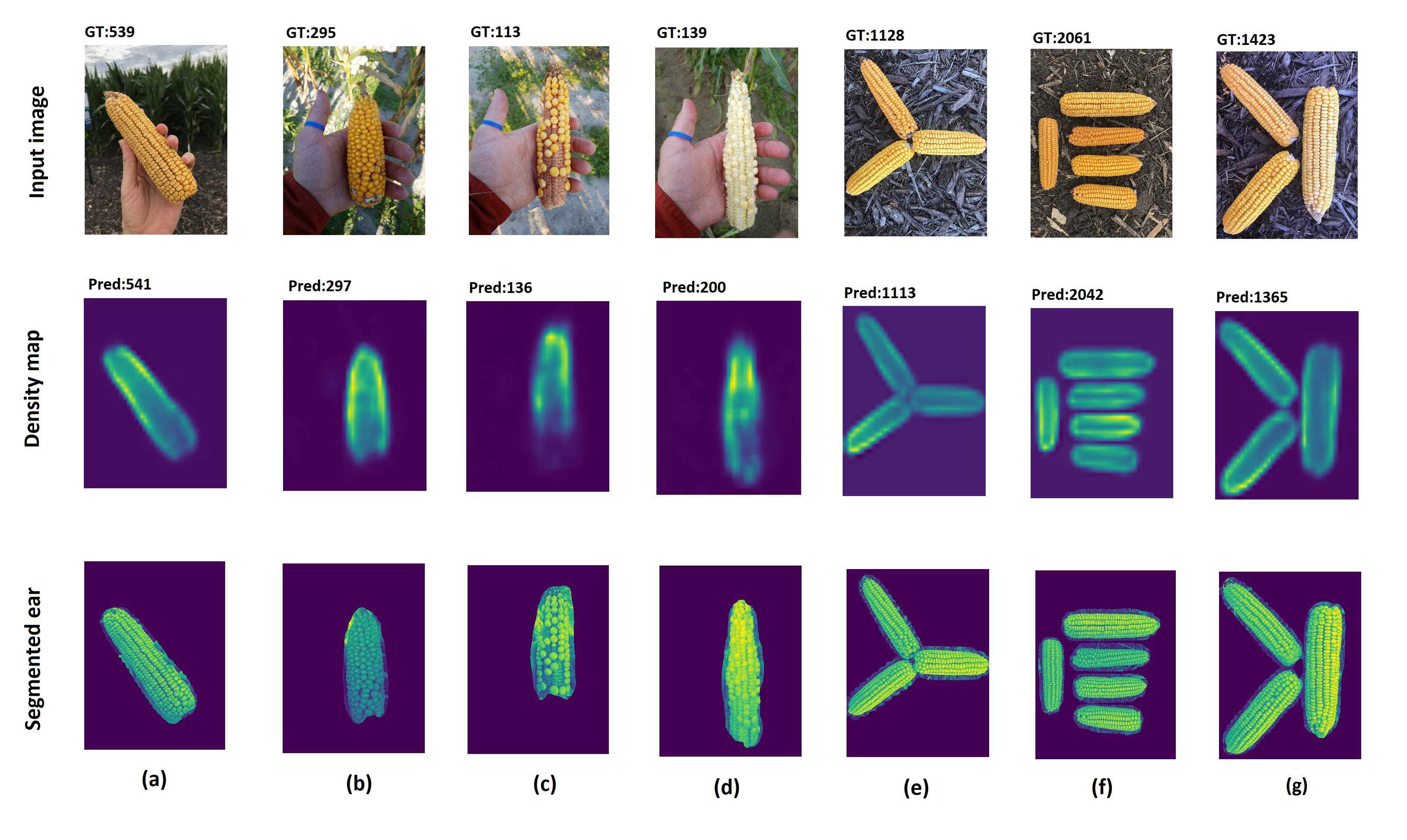}
    \caption{Visual results of our proposed method. The first, second, and third rows are input images, estimated density maps, and segmented ears, respectively. GT and Pred stand for the ground truth number of kernels and predicted number of kernels, respectively.}
    \label{fig:vis_result}
\end{figure}

Figure \ref{fig:vis_result} visualizes a sample of the results that include original images, estimated density maps, and segmented ear images for our proposed method. As shown in Figure \ref{fig:vis_result}, the estimated and the ground truth counts are close and invariant to size, angle, orientation, background, lighting conditions, and the number of ears per image. The prediction accuracy of the proposed model slightly decreased for the completely white ear (image (d) in Figure \ref{fig:vis_result}) which was mainly due to the fact that most of the training data (more than 95\%) of the proposed model only consist of yellow corn ears. As a result, it might make the model confused between the actual kernel and the corn cob itself due to having the same color. This problem can be solved by adding more white color corn ears to the training data. It is worth mentioning that predicting the total number of kernels on both sides of an ear using a 180-degree image is difficult because corn ears are not 100\% symmetric, and, thus, we have non-reducible error in our estimation.

\section{Analysis}

\subsection{Robustness and Sensitivity Analysis}

To estimate the total number of kernels on an ear using a 180-degree image, we count the number of kernels on the one side of ear and then multiply the counted kernels by 2.10 to estimate the total number of kernels on the entire ear. To evaluate the robustness and sensitivity of our proposed method in using a 180-degree image for kernel counting on entire corn ear, we perform the following analysis. For 257 test ears, we take an image of one side of the ear and then flip the ear to the backside and take another image. Similar to our previous section, the ground truth we will be using will be the number of kernels on the entire ear. For this section, the number of test ears changes from 291 ears to 257 ears because only 257 out of 291 test corn ears have the both frontside and backside images of corn ear which can be used in this experiment. We estimate the total number of kernels on a corn ear using the following scenarios:\\

\noindent \textbf{Frontside estimation:} We estimate the total number of kernel using the front-side image of the ear, and then multiply the counted kernels by the empirical coefficient of 2.10 to consider the back side of ear.

\noindent \textbf{Backside estimation:} We rotate the ear 180 degrees and estimate the total number of kernel using the back-side image of the ear. Then, we multiply the counted kernels by the empirical coefficient of 2.10 to also consider the front side of ear.

\noindent \textbf{Both side estimation:} We estimate the total number of kernels on an ear using images of both sides of the ear. As such, the total number of kernels is equal to the sum of the estimated kernels on the front and back sides of the ear.

\begin{figure}[H]
    \centering
    \includegraphics[scale=0.06]{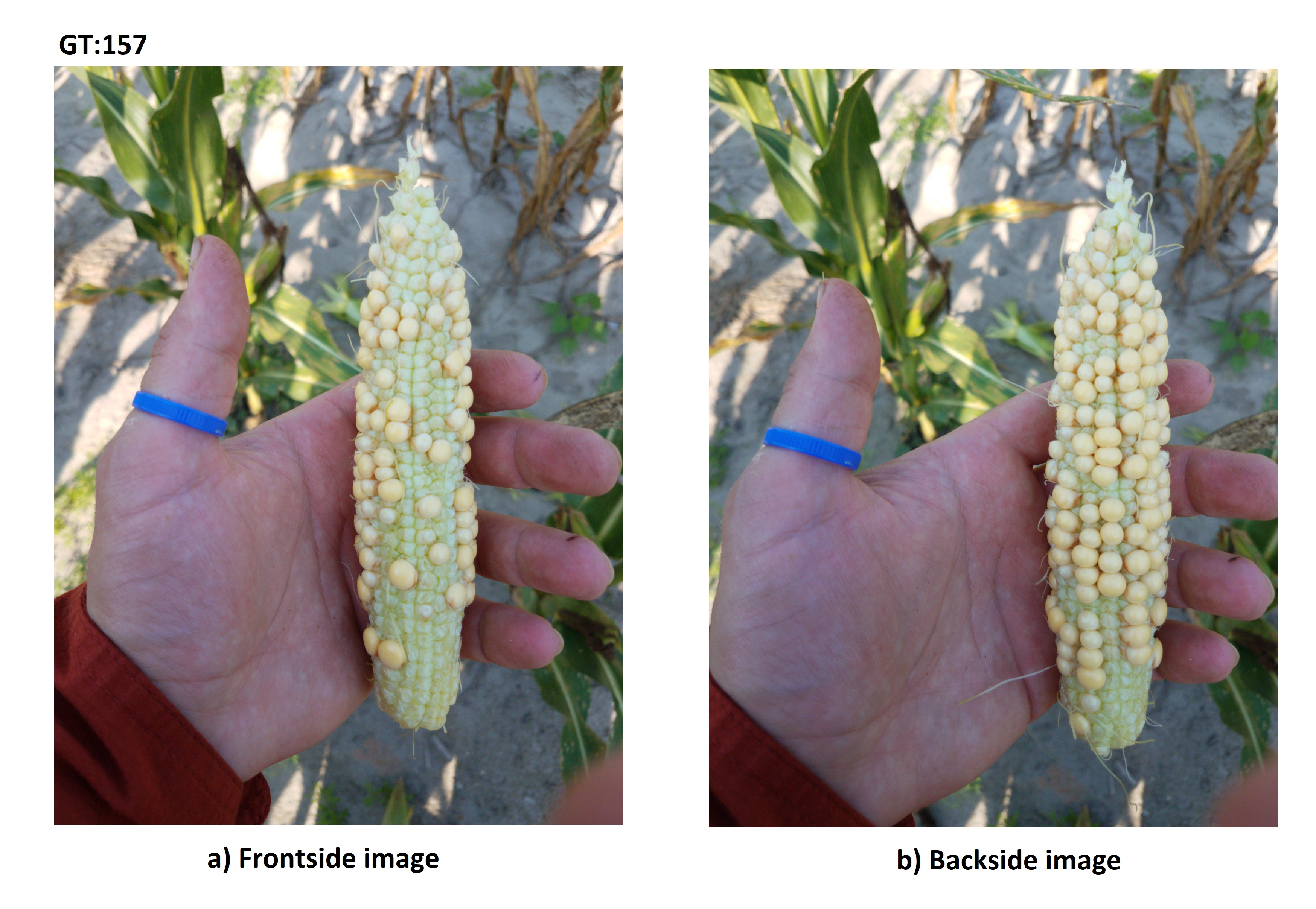}
    \caption{Images (a) and (b) shows the frontside and backside of a test ear with heterogeneous kernel distribution, respectively. The frontside, backside and both side kernel estimation for this ear using DeepCorn method are 144, 227, and 177, respectively. The ground truth number of kernels for this ear is 157.  As a result, we recommend using images of both sides for ears with heterogeneous kernel distribution.}
    \label{fig:two_sides}
\end{figure}

Table \ref{tab:result_rotate} shows the kernel counting performances of our proposed method (DeepCorn) and SaCNN method in our above-mentioned experiment. As shown in Table \ref{tab:result_rotate}, the proposed method outperformed the other method in all three scenarios. The results indicate that our proposed method can estimate the number of kernels with reasonable accuracy using 180-degree images and is robust no matter which side the ear image is captured on. If we compare the frontside and backside performances of the DeepCorn and SaCNN, we see that DeepCorn shows more sensitivity for two main reasons: (1) as shown in Figure \ref{fig:two_sides}, there are some ears in the test dataset for which the density of kernels on the front and back sides are significantly different which makes the estimation of kernels based on only a 180-degree image difficult, and (2) our proposed method is more accurate than the SaCNN method which makes its prediction more sensitive for ears that have different density of kernels on the front and back sides. As shown in Table \ref{tab:result_rotate}, the proposed method is most accurate when using both side images of ears mainly because corn ears are not 100\% symmetric and we recommend using images of both sides for ears with heterogeneous kernel distribution.

\begin{table}[H]
    \centering
    \begin{tabular}{|c|c|c|c|}
    \hline
        Method &  MAE & RMSE &MAPE \\
         \hline
        SaCNN (frontside)  & 50.81 &70.54&54.02\\
            \hline
          DeepCorn (frontside) & \textbf{41.62} & \textbf{60.24}& \textbf{37.40}\\
        \hline

           SaCNN (backside)  & 50.45 &71.26&49.73\\
               \hline
             DeepCorn (backside) & \textbf{44.71} & \textbf{67.07}&\textbf{36.72} \\

        \hline
           SaCNN (bothside)  & 47.61 &66.91&50.15\\
               \hline
             DeepCorn (bothside) & \textbf{33.03} & \textbf{52.11}&\textbf{29.72} \\

        \hline
    \end{tabular}
    \caption{The comparison of the DeepCorn and SaCNN methods on kernel counting performance on the 257 test corn ears. We use images of frontside, backside and both sides of ear for estimation of kernels on the both sides of ear. The DeepCorn is the teacher model version in this experiment.}
    \label{tab:result_rotate}
\end{table}

To further compare the counting performances of the SaCNN and our proposed models, we report the total number of counting errors of these models based on both sides kernel estimation in 257 test corn ears.

\begin{table}[H]
    \centering
    \begin{tabular}{|c|c|c|}
    \hline
      Method  & Miss-counted Kernels &Correctly Counted Kernels \\
         \hline
       SaCNN  & 12,238 &42,379\\
       \hline
       DeepCorn &8,489&46,128\\
       \hline
    \end{tabular}
    \caption{The total number of miss-counted and correctly counted kernels. The total number of all existing kernels in the 257 test corn ears is equal to 54,617.}
    \label{tab:result_subset}
\end{table}

It is worth mentioning that the 257 test corn dataset used in our experiment is considered a difficult test data set for the following reason. Our training data consist of mostly yellow corn ears while the majority of test ears have white colors which exist in some corn breeds and in corn ears harvested early before reaching full maturity. Such white ears can make the model confused between the kernel and the cob itself due to having the same color. This problem can be solved by adding more white ears to the training data. As a result, we report the kernel counting performances of the SaCNN and the proposed model in the subset of test dataset which only includes yellow corn ears. As shown in Table \ref{tab:result_sub_yellow}, the overall prediction error of the proposed model is $9.5\%$ on yellow corn ears which indicates that the miss-counted kernels are mostly attributed to the white corn ears due to the scarcity of such ears in the training data. Table \ref{tab:yellow_part} also shows the performances of our proposed method and the SaCNN in the subset of test dataset which only includes yellow corn ears with respect to all evaluation metrics.

\begin{table}[H]
    \centering
    \begin{tabular}{|c|c|c|}
    \hline
      Method  & Miss-counted Kernels &Correctly Counted Kernels \\
         \hline
       SaCNN  & 1,609 &10,895\\
       \hline
       DeepCorn &1,190&11,314\\
       \hline
    \end{tabular}
    \caption{The total number of miss-counted and correctly counted kernels in the subset of test data which only includes yellow corn ears. The total number of all existing kernels in the 54 yellow test corn ears is equal to 12504.}
    \label{tab:result_sub_yellow}
\end{table}

\begin{table}[H]
    \centering
    \begin{tabular}{|c|c|c|c|}
    \hline
        Method &  MAE & RMSE &MAPE \\
         \hline
        SaCNN (frontside)  & 30.44 &39.66&20.99\\
            \hline
          DeepCorn (frontside) & \textbf{21.19} & \textbf{25.91}& \textbf{11.21}\\
        \hline

           SaCNN (backside)  & 38.52 &\textbf{50.15}&23.52\\
               \hline
             DeepCorn (backside) & \textbf{34.96} & 50.82&\textbf{19.31} \\

        \hline
           SaCNN (bothside)  & 29.80 &37.86&19.33\\
               \hline
             DeepCorn (bothside) & \textbf{22.04} & \textbf{30.29}&\textbf{11.16} \\

        \hline
    \end{tabular}
    \caption{The comparison of the DeepCorn and SaCNN methods on kernel counting performance in the subset of test data which only includes 54 yellow corn ears. We use images of frontside, backside and both sides of ear for estimation of kernels on the both sides of ear. The DeepCorn is the teacher model version in this experiment.}
    \label{tab:yellow_part}
\end{table}

\begin{figure}[H]
    \centering
    \includegraphics[scale=0.28]{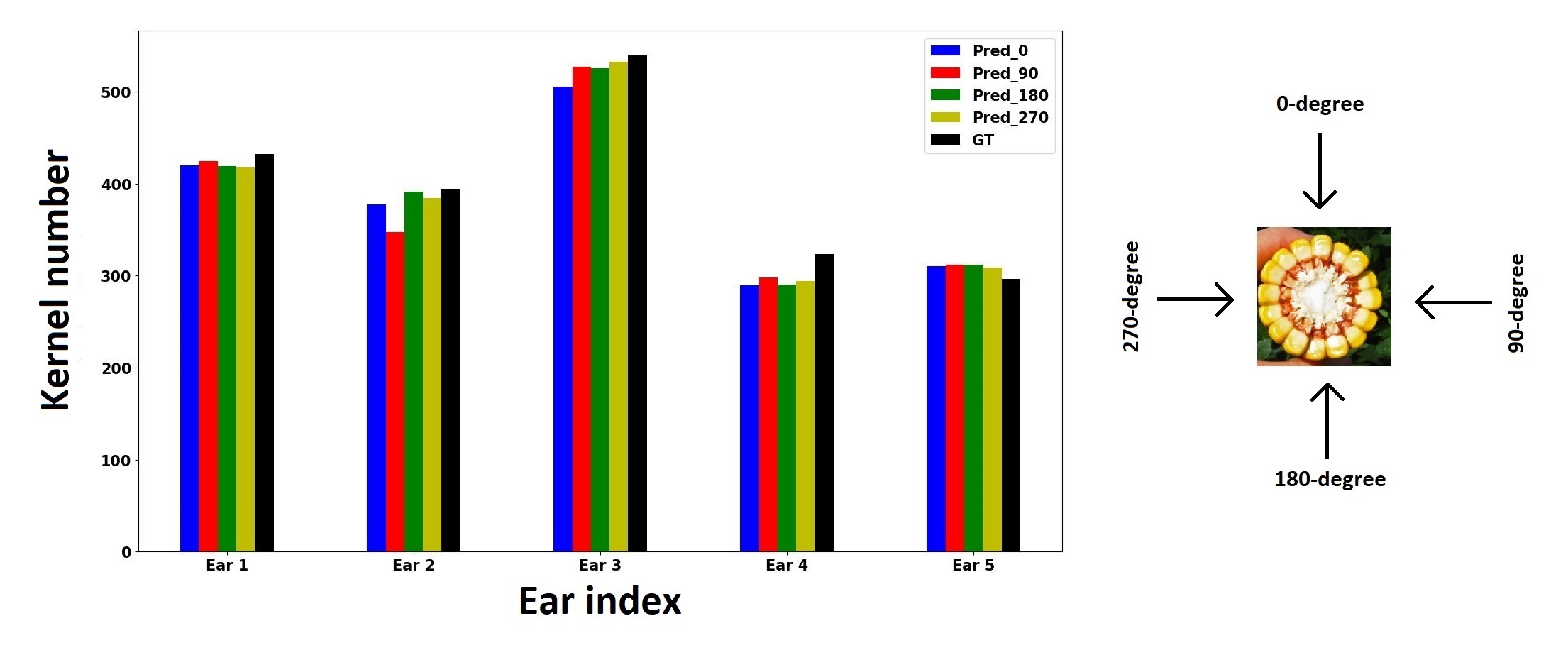}
    \caption{Bar plot of estimated kernels for five ears at four different angels which are 0, 90, 180, and 270 degrees using our proposed method. GT stands for ground truth number of kernels.}\label{fig:barplot}
\end{figure}

To further analyze the effect of kernel distribution on kernel estimation using a single 180-degree image, we use five normal ears with homogeneous kernel distribution and estimate their total number of kernels four times based on images taken at angles 0, 90, 180, 270 degrees using our proposed method. That is, we simply rotate the ear and take an image from each side with the hope that, no matter which side, we achieve a consistent kernel estimation count. Figure \ref{fig:barplot} displays the barplot of estimated kernels for these five ears at four different angels. As shown in Figure \ref{fig:barplot}, the estimations at different angles are close to each other which indicates our proposed method is robust against the image angle for ears with homogeneous kernel distribution.  

\subsection{Yield Estimation}

As previously mentioned, the main application of our corn kernel counting methodology is to estimate in-season corn yield (i.e. before harvest). Having an accurate, efficient yield estimator enables farmers and agronomists to make real-time grain management decisions to maximize yield and minimize potential losses of profitability. This method, which is often called yield component method, enables farmers and agronomists to estimate corn yields accurately within $\pm 20$ bushels per acre \citep{licht2017estimating}.  This procedure requires taking a representative sample of ears from the field and manually counting the kernels to determine the average number of kernels per ear. However, because a healthy ear of corn contains 650 - 800 kernels, manually counting individual kernels is time-consuming, labor-intensive and subject to human error. Additionally, a farmer will need a large amount of ears to achieve an accurate estimation of yield, thus adding to an already time-consuming and exhausting task. 

To remedy this bottleneck, our proposed kernel counting method can be used to count multiple ears in a short time period to accelerate the yield estimation process allowing for real-time in-field yield estimation. \cite{licht2017estimating} provides a classical way to estimate corn yield based on kernel counts. The formula is based on the following components:

\noindent \textbf{Stand counts (plants per acre):} This factor is the number of plants per acre and is usually determined by the number of seed planted, seed quality, and other environmental factors. The number of ears per plant is considered to be one because most corn hybrids have one dominant ear which produces kernels.

\noindent \textbf{Kernel weight:} A kernel can weigh in the range 0.26-0.27 grams. The variation in kernel weight is a response to environmental stress, higher levels of environmental stress will decrease kernel weight, lower levels of stress will have the inverse effect.  Accurately measuring kernel weights is difficult and time consuming, so estimations are used to provide an evaluation of potential yield for farmers and agronomists. Total kernel weight per bushel can range from 65,000 and 100,000 kernels per bushel. Ninety thousand kernels per bushel is the industry default when conducting yield estimations. Outside of extreme weather conditions, kernel weight on average is between 85,000 and 90,000 kernels per bushel \citep{ngoune2020estimation}.

\noindent \textbf{Average number of kernel per ear:} This factor indicates the average number of kernels per ear determined by taking a representative sample of ears from fields and count their kernels. Combining these factors, \cite{licht2017estimating} provides a way to estimate corn yield using Equation \ref{eq:yield_stimation} in bushels per acre. 

\begin{eqnarray}
\textit{Corn Yield}=\frac{\frac{Ears}{Acre}\times \frac{\textit {Average Kernel}}{Ear}}{\frac{\textit{90,000 Kernels}}{ Bushel}}\label{eq:yield_stimation}
\end{eqnarray}

\begin{figure}[H]
    \centering
    \includegraphics[scale=0.15]{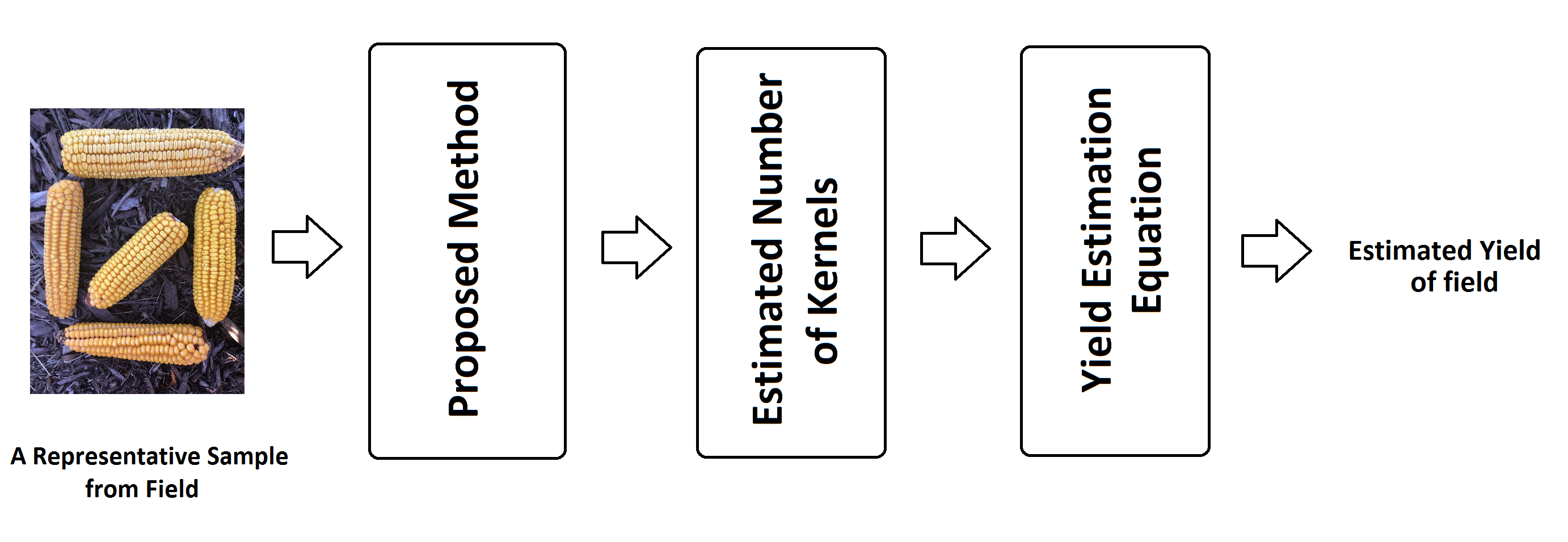}
    \caption{The diagram of the yield estimation procedure.}
    \label{fig:diagram_yield}
\end{figure}

Figure \ref{fig:diagram_yield} shows the diagram of the yield estimation procedure. 

To show how the proposed yield estimation method can be used, we perform the following analysis. Let assume 3 different corn seeds have been planted in 3 different fields and these seeds can be categorized as low, medium, and high yielding which is determined by the size of the produced corn ears. Let also assume 32,000 corn stands have been planted in all 3 fields. Table \ref{tab:yield_estimation} shows the yield estimation results. As shown in Table \ref{tab:yield_estimation}, the size of ear has significant effect on the final estimated yield. To reduce the yield estimation error, we recommend to use a batch ears which represents the overall condition of field well.

\begin{table}[H]
    \centering
    \begin{tabular}{|c|c|c|c|c|}
    \hline
    Input Image     & \begin{tabular}{c}
     Seed\\
     Type
\end{tabular} & 
\begin{tabular}{c}
     Estimated \\
     Kernels
\end{tabular}&  
\begin{tabular}{c}
   Stand\\ Counts 
\end{tabular} & 
\begin{tabular}{c}
  Estimated \\Yield (bu/ac)
\end{tabular}
 \\
     \hline
       
       \raisebox{-\totalheight}{\includegraphics[width=45mm, height=30mm]{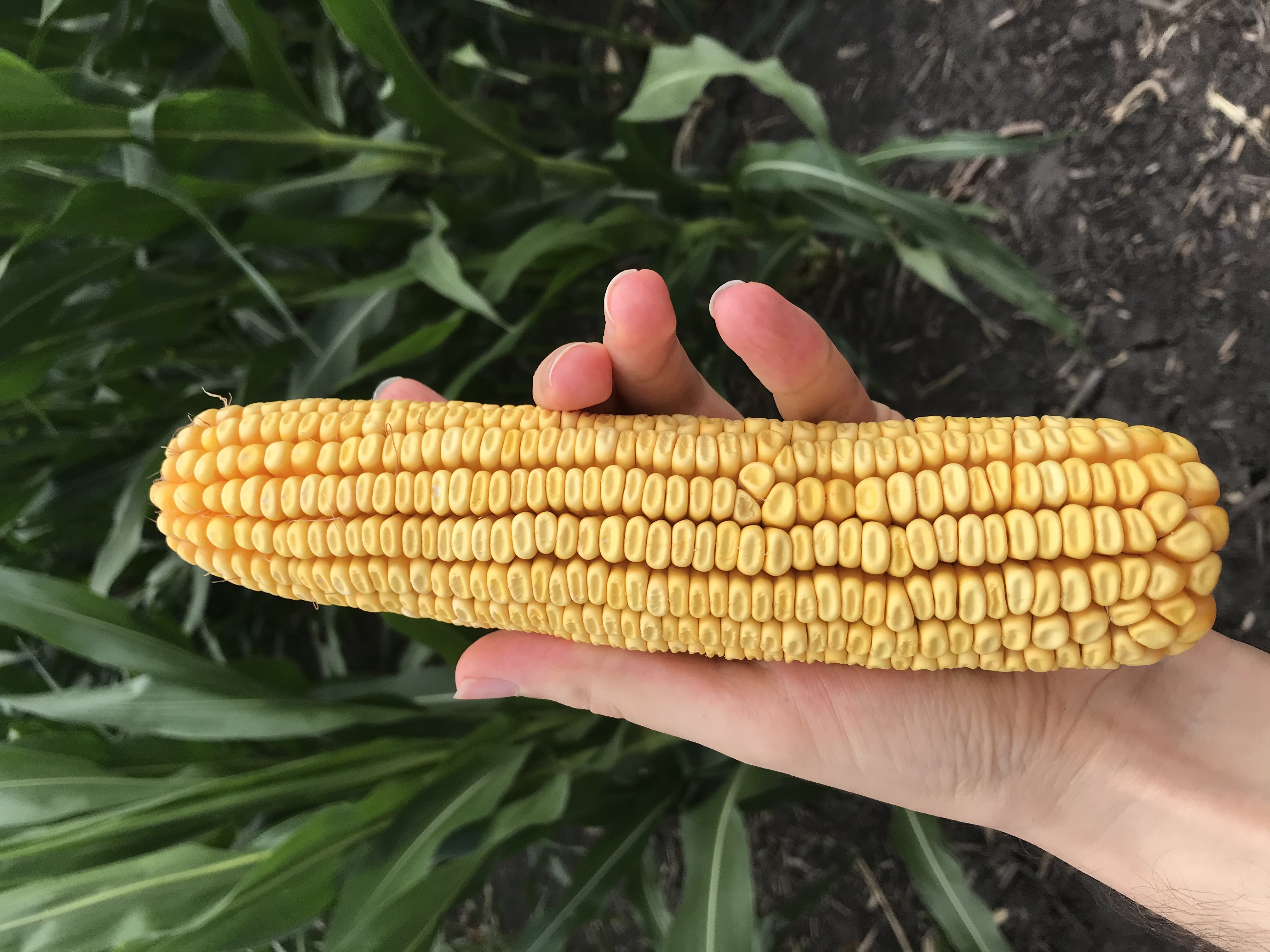}}
   &High Yielding  & 646&32,000&229.69\\
     \hline

       \raisebox{-\totalheight}{\includegraphics[width=45mm, height=30mm]{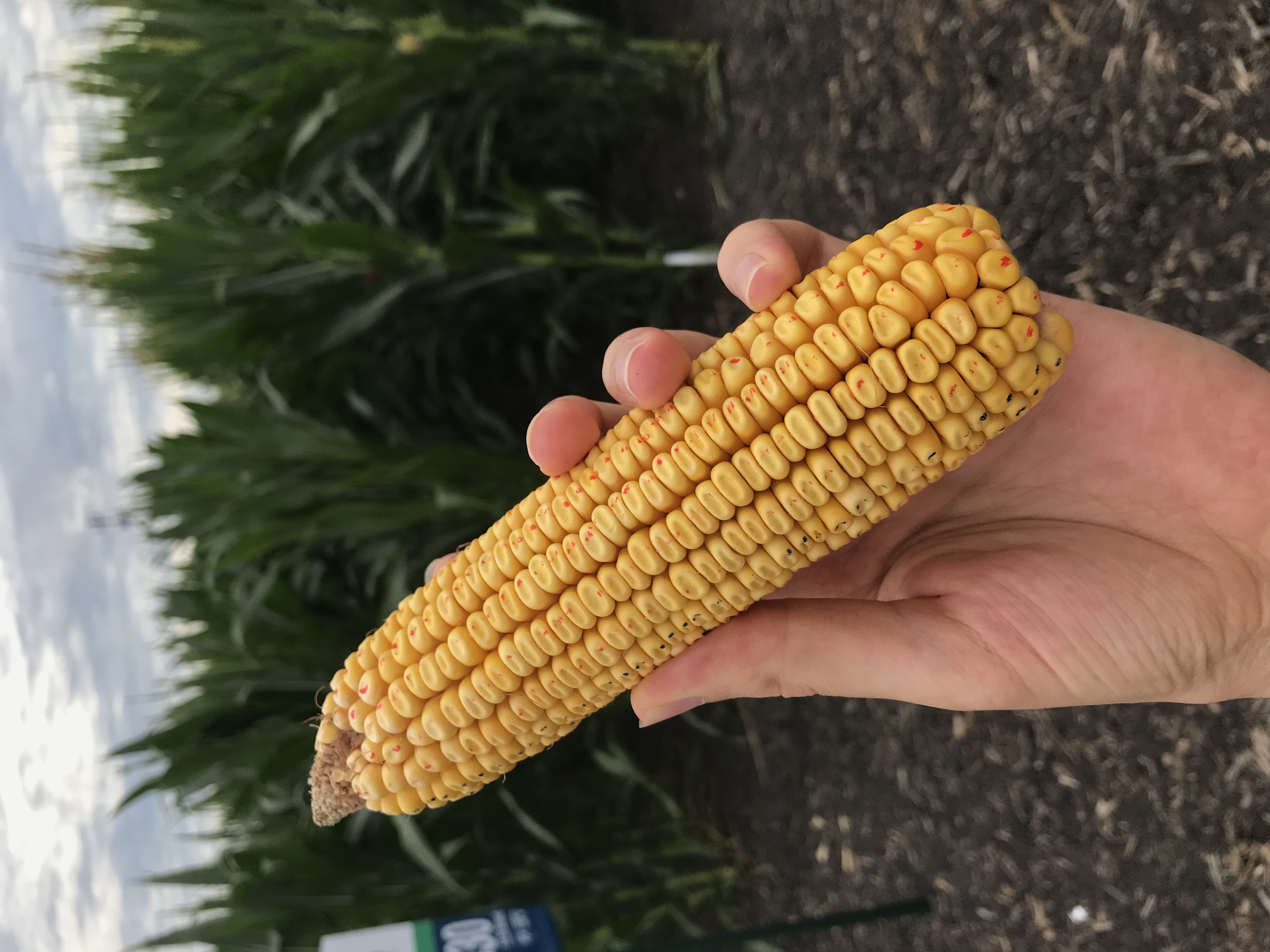}} &Medium Yielding & 541&32,000&192.36\\
     \hline

        \raisebox{-\totalheight}{\includegraphics[width=45mm, height=30mm]{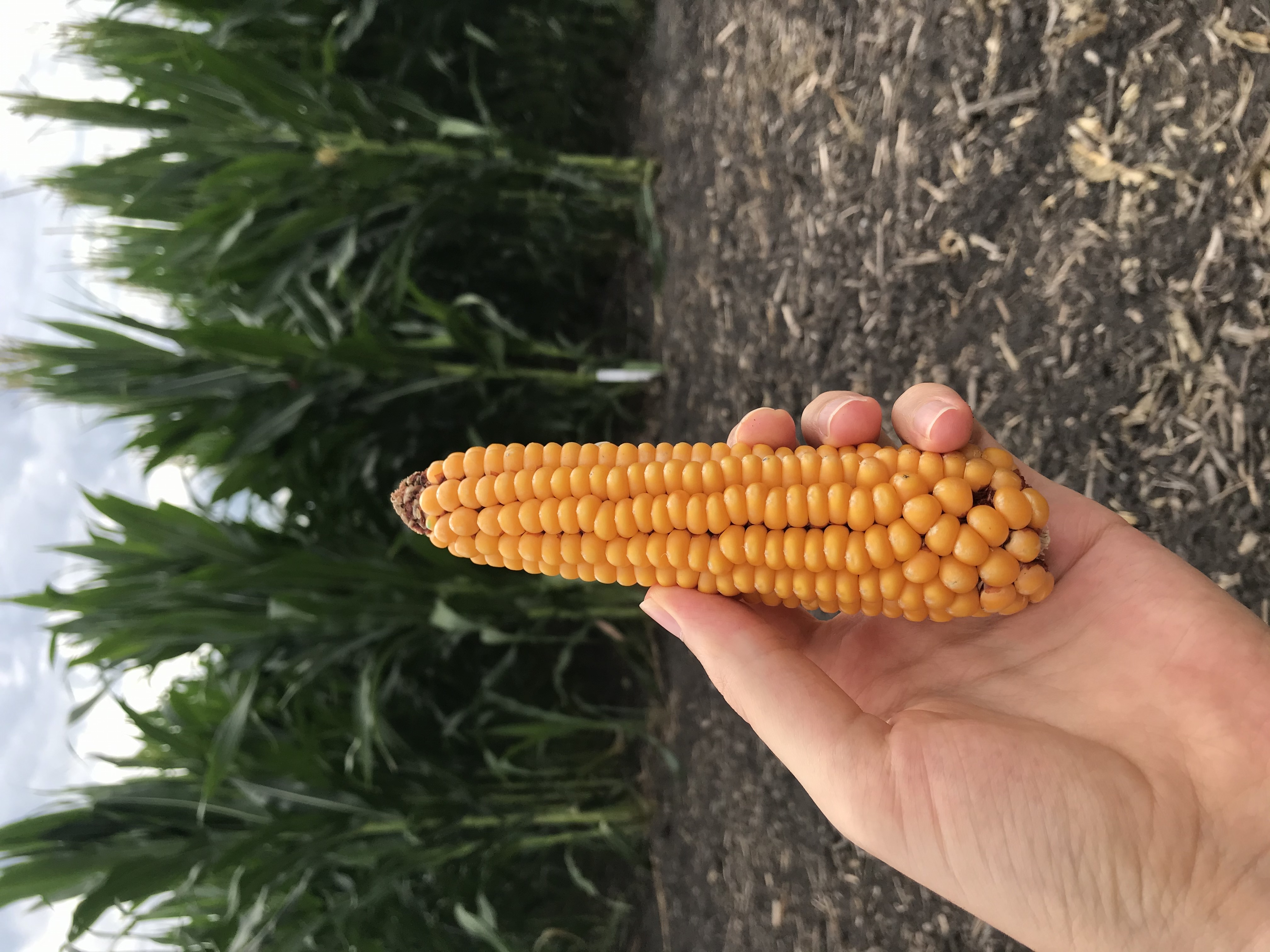}} &Low Yielding  & 300&32,000&106.67\\
     \hline
    
    \end{tabular}
    \caption{The yield estimation results based on 3 different ears. bu/ac stands for bushels per acre.}
    \label{tab:yield_estimation}
\end{table}

\section{Conclusion}

In this paper, we presented a deep learning based method named DeepCorn for corn kernel counting problem. The proposed model uses VGG-16 as backbone for feature extraction. To make the proposed model robust against scale variations, the proposed network merged feature maps from multiple scales of the network using skip connections. In the end, we upsampled the output of the last layer of the network using bilinear interpolation to estimate the density map. To further improve the performance of the proposed method, we used a semi-supervised learning approach to generate pseudo density maps. Then, these pseudo density maps along with the original density maps were used to train our proposed method. Our extensive experimental results demonstrate that our proposed method can successfully count the corn kernels on corn ears regardless of their orientations and the lighting condition as well as out-perform other state of the art methods commonly used in similar tasks. The results also indicated that semi-supervised learning approach improves the accuracy of the proposed method. Our experimental results demonstrated that our proposed method can predict the number of kernels accurately using a 180-degree image and is robust no matter
which side the ear image is captured on. However, if a corn ear is significantly non-symmetric, it is best to estimate the number of kernels using both frontside and backside images of ear.

Our proposed method can be integrated with yield estimation methods to do real-time in-season yield estimation. The yield estimation methods rely on taking a representative sample of ears from the field and manually counting the kernels to determine the average number of kernels per ear. As a result, our proposed corn kernel counting method can be used to count multiple ears in a short time period to accelerate the yield estimation process. We hope that our work leads to the advancement of high throughput phenotyping to benefit plant science as a whole.

\section*{Conflicts of Interest}

The authors declare no conflict of interest.

\section*{Acknowledgement}

This work was partially supported by the National Science Foundation under the LEAP HI and GOALI programs (grant number 1830478) and under the EAGER program (grant number 1842097). Additionally this work was partially supported by Syngenta.

\bibliographystyle{elsarticle-harv}  
\bibliography{references}

\begin{thebibliography}{72}
\expandafter\ifx\csname natexlab\endcsname\relax\def\natexlab#1{#1}\fi
\providecommand{\url}[1]{\texttt{#1}}
\providecommand{\href}[2]{#2}
\providecommand{\path}[1]{#1}
\providecommand{\DOIprefix}{doi:}
\providecommand{\ArXivprefix}{arXiv:}
\providecommand{\URLprefix}{URL: }
\providecommand{\Pubmedprefix}{pmid:}
\providecommand{\doi}[1]{\href{http://dx.doi.org/#1}{\path{#1}}}
\providecommand{\Pubmed}[1]{\href{pmid:#1}{\path{#1}}}
\providecommand{\bibinfo}[2]{#2}
\ifx\xfnm\relax \def\xfnm[#1]{\unskip,\space#1}\fi
%Type = Inproceedings
\bibitem[{Abadi et~al.(2016)Abadi, Barham, Chen, Chen, Davis, Dean, Devin,
  Ghemawat, Irving, Isard et~al.}]{abadi2016tensorflow}
\bibinfo{author}{Abadi, M.}, \bibinfo{author}{Barham, P.},
  \bibinfo{author}{Chen, J.}, \bibinfo{author}{Chen, Z.},
  \bibinfo{author}{Davis, A.}, \bibinfo{author}{Dean, J.},
  \bibinfo{author}{Devin, M.}, \bibinfo{author}{Ghemawat, S.},
  \bibinfo{author}{Irving, G.}, \bibinfo{author}{Isard, M.}, et~al.,
  \bibinfo{year}{2016}.
\newblock \bibinfo{title}{Tensorflow: A system for large-scale machine
  learning}, in: \bibinfo{booktitle}{12th $\{$USENIX$\}$ Symposium on Operating
  Systems Design and Implementation ($\{$OSDI$\}$ 16)}, pp.
  \bibinfo{pages}{265--283}.
%Type = Book
\bibitem[{Bennetzen and Hake(2008)}]{bennetzen2008handbook}
\bibinfo{author}{Bennetzen, J.L.}, \bibinfo{author}{Hake, S.C.},
  \bibinfo{year}{2008}.
\newblock \bibinfo{title}{Handbook of maize: its biology}.
\newblock \bibinfo{publisher}{Springer Science \& Business Media}.
%Type = Article
\bibitem[{Berardi et~al.(2019)Berardi, Hartman, DeLucia and
  Hudiburg}]{berardi2019flooding}
\bibinfo{author}{Berardi, D.}, \bibinfo{author}{Hartman, M.D.},
  \bibinfo{author}{DeLucia, E.H.}, \bibinfo{author}{Hudiburg, T.W.},
  \bibinfo{year}{2019}.
\newblock \bibinfo{title}{Flooding in the us corn belt: Mitigating climate
  change and crop loss by converting to flood tolerant bioenergy crops.}
\newblock \bibinfo{journal}{AGUFM} \bibinfo{volume}{2019},
  \bibinfo{pages}{B33E--04}.
%Type = Inproceedings
\bibitem[{Boominathan et~al.(2016)Boominathan, Kruthiventi and
  Babu}]{boominathan2016crowdnet}
\bibinfo{author}{Boominathan, L.}, \bibinfo{author}{Kruthiventi, S.S.},
  \bibinfo{author}{Babu, R.V.}, \bibinfo{year}{2016}.
\newblock \bibinfo{title}{Crowdnet: A deep convolutional network for dense
  crowd counting}, in: \bibinfo{booktitle}{Proceedings of the 24th ACM
  international conference on Multimedia}, pp. \bibinfo{pages}{640--644}.
%Type = Inproceedings
\bibitem[{Cao et~al.(2018)Cao, Wang, Zhao and Su}]{cao2018scale}
\bibinfo{author}{Cao, X.}, \bibinfo{author}{Wang, Z.}, \bibinfo{author}{Zhao,
  Y.}, \bibinfo{author}{Su, F.}, \bibinfo{year}{2018}.
\newblock \bibinfo{title}{Scale aggregation network for accurate and efficient
  crowd counting}, in: \bibinfo{booktitle}{Proceedings of the European
  Conference on Computer Vision (ECCV)}, pp. \bibinfo{pages}{734--750}.
%Type = Inproceedings
\bibitem[{Caron et~al.(2018)Caron, Bojanowski, Joulin and
  Douze}]{caron2018deep}
\bibinfo{author}{Caron, M.}, \bibinfo{author}{Bojanowski, P.},
  \bibinfo{author}{Joulin, A.}, \bibinfo{author}{Douze, M.},
  \bibinfo{year}{2018}.
\newblock \bibinfo{title}{Deep clustering for unsupervised learning of visual
  features}, in: \bibinfo{booktitle}{Proceedings of the European Conference on
  Computer Vision (ECCV)}, pp. \bibinfo{pages}{132--149}.
%Type = Inproceedings
\bibitem[{Chan et~al.(2008)Chan, Liang and Vasconcelos}]{chan2008privacy}
\bibinfo{author}{Chan, A.B.}, \bibinfo{author}{Liang, Z.S.J.},
  \bibinfo{author}{Vasconcelos, N.}, \bibinfo{year}{2008}.
\newblock \bibinfo{title}{Privacy preserving crowd monitoring: Counting people
  without people models or tracking}, in: \bibinfo{booktitle}{2008 IEEE
  Conference on Computer Vision and Pattern Recognition},
  \bibinfo{organization}{IEEE}. pp. \bibinfo{pages}{1--7}.
%Type = Inproceedings
\bibitem[{Chan and Vasconcelos(2009)}]{chan2009bayesian}
\bibinfo{author}{Chan, A.B.}, \bibinfo{author}{Vasconcelos, N.},
  \bibinfo{year}{2009}.
\newblock \bibinfo{title}{Bayesian poisson regression for crowd counting}, in:
  \bibinfo{booktitle}{2009 IEEE 12th international conference on computer
  vision}, \bibinfo{organization}{IEEE}. pp. \bibinfo{pages}{545--551}.
%Type = Inproceedings
\bibitem[{Chen et~al.(2019)Chen, Bin, Sang and Gao}]{chen2019scale}
\bibinfo{author}{Chen, X.}, \bibinfo{author}{Bin, Y.}, \bibinfo{author}{Sang,
  N.}, \bibinfo{author}{Gao, C.}, \bibinfo{year}{2019}.
\newblock \bibinfo{title}{Scale pyramid network for crowd counting}, in:
  \bibinfo{booktitle}{2019 IEEE Winter Conference on Applications of Computer
  Vision (WACV)}, \bibinfo{organization}{IEEE}. pp.
  \bibinfo{pages}{1941--1950}.
%Type = Article
\bibitem[{Cruz et~al.(2017)Cruz, Luvisi, De~Bellis and Ampatzidis}]{cruz2017x}
\bibinfo{author}{Cruz, A.C.}, \bibinfo{author}{Luvisi, A.},
  \bibinfo{author}{De~Bellis, L.}, \bibinfo{author}{Ampatzidis, Y.},
  \bibinfo{year}{2017}.
\newblock \bibinfo{title}{X-fido: An effective application for detecting olive
  quick decline syndrome with deep learning and data fusion}.
\newblock \bibinfo{journal}{Frontiers in plant science} \bibinfo{volume}{8},
  \bibinfo{pages}{1741}.
%Type = Inproceedings
\bibitem[{Deb and Ventura(2018)}]{deb2018aggregated}
\bibinfo{author}{Deb, D.}, \bibinfo{author}{Ventura, J.}, \bibinfo{year}{2018}.
\newblock \bibinfo{title}{An aggregated multicolumn dilated convolution network
  for perspective-free counting}, in: \bibinfo{booktitle}{Proceedings of the
  IEEE Conference on Computer Vision and Pattern Recognition Workshops}, pp.
  \bibinfo{pages}{195--204}.
%Type = Article
\bibitem[{Furbank and Tester(2011)}]{furbank2011phenomics}
\bibinfo{author}{Furbank, R.T.}, \bibinfo{author}{Tester, M.},
  \bibinfo{year}{2011}.
\newblock \bibinfo{title}{Phenomics--technologies to relieve the phenotyping
  bottleneck}.
\newblock \bibinfo{journal}{Trends in plant science} \bibinfo{volume}{16},
  \bibinfo{pages}{635--644}.
%Type = Article
\bibitem[{Gao et~al.(2020a)Gao, Gao, Liu, Wang and Wang}]{gao2020cnn}
\bibinfo{author}{Gao, G.}, \bibinfo{author}{Gao, J.}, \bibinfo{author}{Liu,
  Q.}, \bibinfo{author}{Wang, Q.}, \bibinfo{author}{Wang, Y.},
  \bibinfo{year}{2020}a.
\newblock \bibinfo{title}{Cnn-based density estimation and crowd counting: A
  survey}.
\newblock \bibinfo{journal}{arXiv preprint arXiv:2003.12783} .
%Type = Inproceedings
\bibitem[{Gao et~al.(2020b)Gao, Liu and Wang}]{gao2020counting}
\bibinfo{author}{Gao, G.}, \bibinfo{author}{Liu, Q.}, \bibinfo{author}{Wang,
  Y.}, \bibinfo{year}{2020}b.
\newblock \bibinfo{title}{Counting dense objects in remote sensing images}, in:
  \bibinfo{booktitle}{ICASSP 2020-2020 IEEE International Conference on
  Acoustics, Speech and Signal Processing (ICASSP)},
  \bibinfo{organization}{IEEE}. pp. \bibinfo{pages}{4137--4141}.
%Type = Article
\bibitem[{Gao et~al.(2019)Gao, Wang and Yuan}]{gao2019scar}
\bibinfo{author}{Gao, J.}, \bibinfo{author}{Wang, Q.}, \bibinfo{author}{Yuan,
  Y.}, \bibinfo{year}{2019}.
\newblock \bibinfo{title}{Scar: Spatial-/channel-wise attention regression
  networks for crowd counting}.
\newblock \bibinfo{journal}{Neurocomputing} \bibinfo{volume}{363},
  \bibinfo{pages}{1--8}.
%Type = Inproceedings
\bibitem[{Girshick(2015)}]{girshick2015fast}
\bibinfo{author}{Girshick, R.}, \bibinfo{year}{2015}.
\newblock \bibinfo{title}{Fast r-cnn}, in: \bibinfo{booktitle}{Proceedings of
  the IEEE international conference on computer vision}, pp.
  \bibinfo{pages}{1440--1448}.
%Type = Article
\bibitem[{Giuffrida et~al.(2018)Giuffrida, Doerner and
  Tsaftaris}]{giuffrida2018pheno}
\bibinfo{author}{Giuffrida, M.V.}, \bibinfo{author}{Doerner, P.},
  \bibinfo{author}{Tsaftaris, S.A.}, \bibinfo{year}{2018}.
\newblock \bibinfo{title}{Pheno-deep counter: a unified and versatile deep
  learning architecture for leaf counting}.
\newblock \bibinfo{journal}{The Plant Journal} \bibinfo{volume}{96},
  \bibinfo{pages}{880--890}.
%Type = Inproceedings
\bibitem[{Glorot and Bengio(2010)}]{glorot2010understanding}
\bibinfo{author}{Glorot, X.}, \bibinfo{author}{Bengio, Y.},
  \bibinfo{year}{2010}.
\newblock \bibinfo{title}{Understanding the difficulty of training deep
  feedforward neural networks}, in: \bibinfo{booktitle}{Proceedings of the
  thirteenth international conference on artificial intelligence and
  statistics}, pp. \bibinfo{pages}{249--256}.
%Type = Article
\bibitem[{Guo et~al.(2018)Guo, Zheng, Potgieter, Diot, Watanabe, Noshita,
  Jordan, Wang, Watson, Ninomiya et~al.}]{guo2018aerial}
\bibinfo{author}{Guo, W.}, \bibinfo{author}{Zheng, B.},
  \bibinfo{author}{Potgieter, A.B.}, \bibinfo{author}{Diot, J.},
  \bibinfo{author}{Watanabe, K.}, \bibinfo{author}{Noshita, K.},
  \bibinfo{author}{Jordan, D.R.}, \bibinfo{author}{Wang, X.},
  \bibinfo{author}{Watson, J.}, \bibinfo{author}{Ninomiya, S.}, et~al.,
  \bibinfo{year}{2018}.
\newblock \bibinfo{title}{Aerial imagery analysis--quantifying appearance and
  number of sorghum heads for applications in breeding and agronomy}.
\newblock \bibinfo{journal}{Frontiers in plant science} \bibinfo{volume}{9},
  \bibinfo{pages}{1544}.
%Type = Inproceedings
\bibitem[{Haug and Ostermann(2014)}]{haug2014crop}
\bibinfo{author}{Haug, S.}, \bibinfo{author}{Ostermann, J.},
  \bibinfo{year}{2014}.
\newblock \bibinfo{title}{A crop/weed field image dataset for the evaluation of
  computer vision based precision agriculture tasks}, in:
  \bibinfo{booktitle}{European Conference on Computer Vision},
  \bibinfo{organization}{Springer}. pp. \bibinfo{pages}{105--116}.
%Type = Misc
\bibitem[{He et~al.(2015)He, Zhang, Ren and Sun}]{he2015deep}
\bibinfo{author}{He, K.}, \bibinfo{author}{Zhang, X.}, \bibinfo{author}{Ren,
  S.}, \bibinfo{author}{Sun, J.}, \bibinfo{year}{2015}.
\newblock \bibinfo{title}{Deep residual learning for image recognition}.
\newblock \href{http://arxiv.org/abs/1512.03385}{{\tt arXiv:1512.03385}}.
%Type = Inproceedings
\bibitem[{Idrees et~al.(2013)Idrees, Saleemi, Seibert and
  Shah}]{idrees2013multi}
\bibinfo{author}{Idrees, H.}, \bibinfo{author}{Saleemi, I.},
  \bibinfo{author}{Seibert, C.}, \bibinfo{author}{Shah, M.},
  \bibinfo{year}{2013}.
\newblock \bibinfo{title}{Multi-source multi-scale counting in extremely dense
  crowd images}, in: \bibinfo{booktitle}{Proceedings of the IEEE conference on
  computer vision and pattern recognition}, pp. \bibinfo{pages}{2547--2554}.
%Type = Article
\bibitem[{Jiang et~al.(2020a)Jiang, Zhang, Zhang, Lv, Zhou, Pang, Xu and
  Xu}]{jiang2020density}
\bibinfo{author}{Jiang, X.}, \bibinfo{author}{Zhang, L.},
  \bibinfo{author}{Zhang, T.}, \bibinfo{author}{Lv, P.}, \bibinfo{author}{Zhou,
  B.}, \bibinfo{author}{Pang, Y.}, \bibinfo{author}{Xu, M.},
  \bibinfo{author}{Xu, C.}, \bibinfo{year}{2020}a.
\newblock \bibinfo{title}{Density-aware multi-task learning for crowd
  counting}.
\newblock \bibinfo{journal}{IEEE Transactions on Multimedia} .
%Type = Article
\bibitem[{Jiang et~al.(2020b)Jiang, Li et~al.}]{jiang2020convolutional}
\bibinfo{author}{Jiang, Y.}, \bibinfo{author}{Li, C.}, et~al.,
  \bibinfo{year}{2020}b.
\newblock \bibinfo{title}{Convolutional neural networks for image-based
  high-throughput plant phenotyping: A review}.
\newblock \bibinfo{journal}{Plant Phenomics} \bibinfo{volume}{2020},
  \bibinfo{pages}{4152816}.
%Type = Article
\bibitem[{Khaki et~al.(2019a)Khaki, Khalilzadeh and
  Wang}]{khaki2019classification}
\bibinfo{author}{Khaki, S.}, \bibinfo{author}{Khalilzadeh, Z.},
  \bibinfo{author}{Wang, L.}, \bibinfo{year}{2019}a.
\newblock \bibinfo{title}{Classification of crop tolerance to heat and
  drought—a deep convolutional neural networks approach}.
\newblock \bibinfo{journal}{Agronomy} \bibinfo{volume}{9},
  \bibinfo{pages}{833}.
%Type = Article
\bibitem[{Khaki et~al.(2020a)Khaki, Khalilzadeh and Wang}]{khaki2020predicting}
\bibinfo{author}{Khaki, S.}, \bibinfo{author}{Khalilzadeh, Z.},
  \bibinfo{author}{Wang, L.}, \bibinfo{year}{2020}a.
\newblock \bibinfo{title}{Predicting yield performance of parents in plant
  breeding: A neural collaborative filtering approach}.
\newblock \bibinfo{journal}{Plos one} \bibinfo{volume}{15},
  \bibinfo{pages}{e0233382}.
%Type = Article
\bibitem[{Khaki et~al.(2020b)Khaki, Pham, Han, Kuhl, Kent and
  Wang}]{khaki2020convolutional}
\bibinfo{author}{Khaki, S.}, \bibinfo{author}{Pham, H.}, \bibinfo{author}{Han,
  Y.}, \bibinfo{author}{Kuhl, A.}, \bibinfo{author}{Kent, W.},
  \bibinfo{author}{Wang, L.}, \bibinfo{year}{2020}b.
\newblock \bibinfo{title}{Convolutional neural networks for image-based corn
  kernel detection and counting}.
\newblock \bibinfo{journal}{Sensors} \bibinfo{volume}{20},
  \bibinfo{pages}{2721}.
%Type = Article
\bibitem[{Khaki and Wang(2019)}]{khaki2019crop}
\bibinfo{author}{Khaki, S.}, \bibinfo{author}{Wang, L.}, \bibinfo{year}{2019}.
\newblock \bibinfo{title}{Crop yield prediction using deep neural networks}.
\newblock \bibinfo{journal}{Frontiers in Plant Science} \bibinfo{volume}{10},
  \bibinfo{pages}{621}.
%Type = Article
\bibitem[{Khaki et~al.(2019b)Khaki, Wang and Archontoulis}]{khaki2019cnn}
\bibinfo{author}{Khaki, S.}, \bibinfo{author}{Wang, L.},
  \bibinfo{author}{Archontoulis, S.V.}, \bibinfo{year}{2019}b.
\newblock \bibinfo{title}{A cnn-rnn framework for crop yield prediction}.
\newblock \bibinfo{journal}{Frontiers in Plant Science} \bibinfo{volume}{10}.
%Type = Misc
\bibitem[{Kingma and Ba(2014)}]{kingma2014adam}
\bibinfo{author}{Kingma, D.P.}, \bibinfo{author}{Ba, J.}, \bibinfo{year}{2014}.
\newblock \bibinfo{title}{Adam: A method for stochastic optimization}.
\newblock \href{http://arxiv.org/abs/1412.6980}{{\tt arXiv:1412.6980}}.
%Type = Article
\bibitem[{Krizhevsky et~al.(2017)Krizhevsky, Sutskever and
  Hinton}]{krizhevsky2017imagenet}
\bibinfo{author}{Krizhevsky, A.}, \bibinfo{author}{Sutskever, I.},
  \bibinfo{author}{Hinton, G.E.}, \bibinfo{year}{2017}.
\newblock \bibinfo{title}{Imagenet classification with deep convolutional
  neural networks}.
\newblock \bibinfo{journal}{Communications of the ACM} \bibinfo{volume}{60},
  \bibinfo{pages}{84--90}.
%Type = Misc
\bibitem[{Kumar et~al.(2019)Kumar, Jain, Tripathi, Singh and
  Krishna}]{kumar2019mtcnet}
\bibinfo{author}{Kumar, A.}, \bibinfo{author}{Jain, N.},
  \bibinfo{author}{Tripathi, S.}, \bibinfo{author}{Singh, C.},
  \bibinfo{author}{Krishna, K.}, \bibinfo{year}{2019}.
\newblock \bibinfo{title}{Mtcnet: Multi-task learning paradigm for crowd count
  estimation}.
\newblock \href{http://arxiv.org/abs/1908.08652}{{\tt arXiv:1908.08652}}.
%Type = Article
\bibitem[{LeCun et~al.(1998)LeCun, Bottou, Bengio and
  Haffner}]{lecun1998gradient}
\bibinfo{author}{LeCun, Y.}, \bibinfo{author}{Bottou, L.},
  \bibinfo{author}{Bengio, Y.}, \bibinfo{author}{Haffner, P.},
  \bibinfo{year}{1998}.
\newblock \bibinfo{title}{Gradient-based learning applied to document
  recognition}.
\newblock \bibinfo{journal}{Proceedings of the IEEE} \bibinfo{volume}{86},
  \bibinfo{pages}{2278--2324}.
%Type = Inproceedings
\bibitem[{Li et~al.(2018)Li, Zhang and Chen}]{li2018csrnet}
\bibinfo{author}{Li, Y.}, \bibinfo{author}{Zhang, X.}, \bibinfo{author}{Chen,
  D.}, \bibinfo{year}{2018}.
\newblock \bibinfo{title}{Csrnet: Dilated convolutional neural networks for
  understanding the highly congested scenes}, in:
  \bibinfo{booktitle}{Proceedings of the IEEE conference on computer vision and
  pattern recognition}, pp. \bibinfo{pages}{1091--1100}.
%Type = Article
\bibitem[{Licht(2017)}]{licht2017estimating}
\bibinfo{author}{Licht, M.A.}, \bibinfo{year}{2017}.
\newblock \bibinfo{title}{Estimating corn yields using yield components} .
%Type = Inproceedings
\bibitem[{Liu et~al.(2020)Liu, Chen, Wu, Chen, Li and Lin}]{liu2020efficient}
\bibinfo{author}{Liu, L.}, \bibinfo{author}{Chen, J.}, \bibinfo{author}{Wu,
  H.}, \bibinfo{author}{Chen, T.}, \bibinfo{author}{Li, G.},
  \bibinfo{author}{Lin, L.}, \bibinfo{year}{2020}.
\newblock \bibinfo{title}{Efficient crowd counting via structured knowledge
  transfer}, in: \bibinfo{booktitle}{Proceedings of the 28th ACM International
  Conference on Multimedia}, pp. \bibinfo{pages}{2645--2654}.
%Type = Inproceedings
\bibitem[{Liu et~al.(2019a)Liu, Qiu, Li, Liu, Ouyang and Lin}]{liu2019crowd}
\bibinfo{author}{Liu, L.}, \bibinfo{author}{Qiu, Z.}, \bibinfo{author}{Li, G.},
  \bibinfo{author}{Liu, S.}, \bibinfo{author}{Ouyang, W.},
  \bibinfo{author}{Lin, L.}, \bibinfo{year}{2019}a.
\newblock \bibinfo{title}{Crowd counting with deep structured scale integration
  network}, in: \bibinfo{booktitle}{Proceedings of the IEEE/CVF International
  Conference on Computer Vision}, pp. \bibinfo{pages}{1774--1783}.
%Type = Inproceedings
\bibitem[{Liu et~al.(2018)Liu, Wang, Li, Ouyang and Lin}]{liu2018crowd}
\bibinfo{author}{Liu, L.}, \bibinfo{author}{Wang, H.}, \bibinfo{author}{Li,
  G.}, \bibinfo{author}{Ouyang, W.}, \bibinfo{author}{Lin, L.},
  \bibinfo{year}{2018}.
\newblock \bibinfo{title}{Crowd counting using deep recurrent spatial-aware
  network}, in: \bibinfo{booktitle}{Proceedings of the 27th International Joint
  Conference on Artificial Intelligence}, pp. \bibinfo{pages}{849--855}.
%Type = Inproceedings
\bibitem[{Liu et~al.(2016)Liu, Anguelov, Erhan, Szegedy, Reed, Fu and
  Berg}]{liu2016ssd}
\bibinfo{author}{Liu, W.}, \bibinfo{author}{Anguelov, D.},
  \bibinfo{author}{Erhan, D.}, \bibinfo{author}{Szegedy, C.},
  \bibinfo{author}{Reed, S.}, \bibinfo{author}{Fu, C.Y.},
  \bibinfo{author}{Berg, A.C.}, \bibinfo{year}{2016}.
\newblock \bibinfo{title}{Ssd: Single shot multibox detector}, in:
  \bibinfo{booktitle}{European conference on computer vision},
  \bibinfo{organization}{Springer}. pp. \bibinfo{pages}{21--37}.
%Type = Inproceedings
\bibitem[{Liu et~al.(2019b)Liu, Salzmann and Fua}]{liu2019context}
\bibinfo{author}{Liu, W.}, \bibinfo{author}{Salzmann, M.},
  \bibinfo{author}{Fua, P.}, \bibinfo{year}{2019}b.
\newblock \bibinfo{title}{Context-aware crowd counting}, in:
  \bibinfo{booktitle}{Proceedings of the IEEE/CVF Conference on Computer Vision
  and Pattern Recognition}, pp. \bibinfo{pages}{5099--5108}.
%Type = Article
\bibitem[{Lu et~al.(2017)Lu, Cao, Xiao, Zhuang and Shen}]{lu2017tasselnet}
\bibinfo{author}{Lu, H.}, \bibinfo{author}{Cao, Z.}, \bibinfo{author}{Xiao,
  Y.}, \bibinfo{author}{Zhuang, B.}, \bibinfo{author}{Shen, C.},
  \bibinfo{year}{2017}.
\newblock \bibinfo{title}{Tasselnet: counting maize tassels in the wild via
  local counts regression network}.
\newblock \bibinfo{journal}{Plant methods} \bibinfo{volume}{13},
  \bibinfo{pages}{79}.
%Type = Article
\bibitem[{Ma et~al.(2019a)Ma, Dai and Tan}]{ma2019atrous}
\bibinfo{author}{Ma, J.}, \bibinfo{author}{Dai, Y.}, \bibinfo{author}{Tan,
  Y.P.}, \bibinfo{year}{2019}a.
\newblock \bibinfo{title}{Atrous convolutions spatial pyramid network for crowd
  counting and density estimation}.
\newblock \bibinfo{journal}{Neurocomputing} \bibinfo{volume}{350},
  \bibinfo{pages}{91--101}.
%Type = Inproceedings
\bibitem[{Ma et~al.(2019b)Ma, Wei, Hong and Gong}]{ma2019bayesian}
\bibinfo{author}{Ma, Z.}, \bibinfo{author}{Wei, X.}, \bibinfo{author}{Hong,
  X.}, \bibinfo{author}{Gong, Y.}, \bibinfo{year}{2019}b.
\newblock \bibinfo{title}{Bayesian loss for crowd count estimation with point
  supervision}, in: \bibinfo{booktitle}{Proceedings of the IEEE/CVF
  International Conference on Computer Vision}, pp.
  \bibinfo{pages}{6142--6151}.
%Type = Article
\bibitem[{Mohanty et~al.(2016)Mohanty, Hughes and
  Salath{\'e}}]{mohanty2016using}
\bibinfo{author}{Mohanty, S.P.}, \bibinfo{author}{Hughes, D.P.},
  \bibinfo{author}{Salath{\'e}, M.}, \bibinfo{year}{2016}.
\newblock \bibinfo{title}{Using deep learning for image-based plant disease
  detection}.
\newblock \bibinfo{journal}{Frontiers in plant science} \bibinfo{volume}{7},
  \bibinfo{pages}{1419}.
%Type = Article
\bibitem[{Mosley et~al.(2020)Mosley, Pham, Bansal and Hare}]{mosley2020image}
\bibinfo{author}{Mosley, L.}, \bibinfo{author}{Pham, H.},
  \bibinfo{author}{Bansal, Y.}, \bibinfo{author}{Hare, E.},
  \bibinfo{year}{2020}.
\newblock \bibinfo{title}{Image-based sorghum head counting when you only look
  once}.
\newblock \bibinfo{journal}{arXiv preprint arXiv:2009.11929} .
%Type = Article
\bibitem[{Naik et~al.(2017)Naik, Zhang, Lofquist, Assefa, Sarkar, Ackerman,
  Singh, Singh and Ganapathysubramanian}]{naik2017real}
\bibinfo{author}{Naik, H.S.}, \bibinfo{author}{Zhang, J.},
  \bibinfo{author}{Lofquist, A.}, \bibinfo{author}{Assefa, T.},
  \bibinfo{author}{Sarkar, S.}, \bibinfo{author}{Ackerman, D.},
  \bibinfo{author}{Singh, A.}, \bibinfo{author}{Singh, A.K.},
  \bibinfo{author}{Ganapathysubramanian, B.}, \bibinfo{year}{2017}.
\newblock \bibinfo{title}{A real-time phenotyping framework using machine
  learning for plant stress severity rating in soybean}.
\newblock \bibinfo{journal}{Plant methods} \bibinfo{volume}{13},
  \bibinfo{pages}{23}.
%Type = Article
\bibitem[{Nazli et~al.(2018)Nazli, Halim, Abdullah, Hussin and
  Samsudin}]{nazli2018potential}
\bibinfo{author}{Nazli, M.H.}, \bibinfo{author}{Halim, R.A.},
  \bibinfo{author}{Abdullah, A.M.}, \bibinfo{author}{Hussin, G.},
  \bibinfo{author}{Samsudin, A.A.}, \bibinfo{year}{2018}.
\newblock \bibinfo{title}{Potential of feeding beef cattle with whole corn crop
  silage and rice straw in malaysia}.
\newblock \bibinfo{journal}{Tropical animal health and production}
  \bibinfo{volume}{50}, \bibinfo{pages}{1119--1124}.
%Type = Article
\bibitem[{Ngoune~Tandzi and Mutengwa(2020)}]{ngoune2020estimation}
\bibinfo{author}{Ngoune~Tandzi, L.}, \bibinfo{author}{Mutengwa, C.S.},
  \bibinfo{year}{2020}.
\newblock \bibinfo{title}{Estimation of maize (zea mays l.) yield per harvest
  area: appropriate methods}.
\newblock \bibinfo{journal}{Agronomy} \bibinfo{volume}{10},
  \bibinfo{pages}{29}.
%Type = Inproceedings
\bibitem[{Redmon et~al.(2016)Redmon, Divvala, Girshick and
  Farhadi}]{redmon2016you}
\bibinfo{author}{Redmon, J.}, \bibinfo{author}{Divvala, S.},
  \bibinfo{author}{Girshick, R.}, \bibinfo{author}{Farhadi, A.},
  \bibinfo{year}{2016}.
\newblock \bibinfo{title}{You only look once: Unified, real-time object
  detection}, in: \bibinfo{booktitle}{Proceedings of the IEEE conference on
  computer vision and pattern recognition}, pp. \bibinfo{pages}{779--788}.
%Type = Article
\bibitem[{Russello(2018)}]{russello2018convolutional}
\bibinfo{author}{Russello, H.}, \bibinfo{year}{2018}.
\newblock \bibinfo{title}{Convolutional neural networks for crop yield
  prediction using satellite images}.
\newblock \bibinfo{journal}{IBM Center for Advanced Studies} .
%Type = Inproceedings
\bibitem[{Sam et~al.(2017)Sam, Surya and Babu}]{sam2017switching}
\bibinfo{author}{Sam, D.B.}, \bibinfo{author}{Surya, S.},
  \bibinfo{author}{Babu, R.V.}, \bibinfo{year}{2017}.
\newblock \bibinfo{title}{Switching convolutional neural network for crowd
  counting}, in: \bibinfo{booktitle}{2017 IEEE Conference on Computer Vision
  and Pattern Recognition (CVPR)}, \bibinfo{organization}{IEEE}. pp.
  \bibinfo{pages}{4031--4039}.
%Type = Article
\bibitem[{Sang et~al.(2019)Sang, Wu, Luo, Xiang, Zhang, Hu and
  Xia}]{sang2019improved}
\bibinfo{author}{Sang, J.}, \bibinfo{author}{Wu, W.}, \bibinfo{author}{Luo,
  H.}, \bibinfo{author}{Xiang, H.}, \bibinfo{author}{Zhang, Q.},
  \bibinfo{author}{Hu, H.}, \bibinfo{author}{Xia, X.}, \bibinfo{year}{2019}.
\newblock \bibinfo{title}{Improved crowd counting method based on
  scale-adaptive convolutional neural network}.
\newblock \bibinfo{journal}{IEEE Access} \bibinfo{volume}{7},
  \bibinfo{pages}{24411--24419}.
%Type = Article
\bibitem[{Shi et~al.(2018)Shi, Zhang, Sun and Ye}]{shi2018multiscale}
\bibinfo{author}{Shi, Z.}, \bibinfo{author}{Zhang, L.}, \bibinfo{author}{Sun,
  Y.}, \bibinfo{author}{Ye, Y.}, \bibinfo{year}{2018}.
\newblock \bibinfo{title}{Multiscale multitask deep netvlad for crowd
  counting}.
\newblock \bibinfo{journal}{IEEE Transactions on Industrial Informatics}
  \bibinfo{volume}{14}, \bibinfo{pages}{4953--4962}.
%Type = Misc
\bibitem[{Simonyan and Zisserman(2014)}]{simonyan2014deep}
\bibinfo{author}{Simonyan, K.}, \bibinfo{author}{Zisserman, A.},
  \bibinfo{year}{2014}.
\newblock \bibinfo{title}{Very deep convolutional networks for large-scale
  image recognition}.
\newblock \href{http://arxiv.org/abs/1409.1556}{{\tt arXiv:1409.1556}}.
%Type = Article
\bibitem[{Singh et~al.(2016)Singh, Ganapathysubramanian, Singh and
  Sarkar}]{singh2016machine}
\bibinfo{author}{Singh, A.}, \bibinfo{author}{Ganapathysubramanian, B.},
  \bibinfo{author}{Singh, A.K.}, \bibinfo{author}{Sarkar, S.},
  \bibinfo{year}{2016}.
\newblock \bibinfo{title}{Machine learning for high-throughput stress
  phenotyping in plants}.
\newblock \bibinfo{journal}{Trends in plant science} \bibinfo{volume}{21},
  \bibinfo{pages}{110--124}.
%Type = Article
\bibitem[{Stephenson et~al.(2010)Stephenson, Newman and
  Mayhew}]{stephenson2010population}
\bibinfo{author}{Stephenson, J.}, \bibinfo{author}{Newman, K.},
  \bibinfo{author}{Mayhew, S.}, \bibinfo{year}{2010}.
\newblock \bibinfo{title}{Population dynamics and climate change: what are the
  links?}
\newblock \bibinfo{journal}{Journal of Public Health} \bibinfo{volume}{32},
  \bibinfo{pages}{150--156}.
%Type = Article
\bibitem[{Sudars et~al.(2020)Sudars, Jasko, Namatevs, Ozola and
  Badaukis}]{sudars2020dataset}
\bibinfo{author}{Sudars, K.}, \bibinfo{author}{Jasko, J.},
  \bibinfo{author}{Namatevs, I.}, \bibinfo{author}{Ozola, L.},
  \bibinfo{author}{Badaukis, N.}, \bibinfo{year}{2020}.
\newblock \bibinfo{title}{Dataset of annotated food crops and weed images for
  robotic computer vision control}.
\newblock \bibinfo{journal}{Data in Brief} , \bibinfo{pages}{105833}.
%Type = Inproceedings
\bibitem[{Szegedy et~al.(2015)Szegedy, Liu, Jia, Sermanet, Reed, Anguelov,
  Erhan, Vanhoucke and Rabinovich}]{szegedy2015going}
\bibinfo{author}{Szegedy, C.}, \bibinfo{author}{Liu, W.}, \bibinfo{author}{Jia,
  Y.}, \bibinfo{author}{Sermanet, P.}, \bibinfo{author}{Reed, S.},
  \bibinfo{author}{Anguelov, D.}, \bibinfo{author}{Erhan, D.},
  \bibinfo{author}{Vanhoucke, V.}, \bibinfo{author}{Rabinovich, A.},
  \bibinfo{year}{2015}.
\newblock \bibinfo{title}{Going deeper with convolutions}, in:
  \bibinfo{booktitle}{Proceedings of the IEEE conference on computer vision and
  pattern recognition}, pp. \bibinfo{pages}{1--9}.
%Type = Article
\bibitem[{Tan and Le(2019)}]{tan2019efficientnet}
\bibinfo{author}{Tan, M.}, \bibinfo{author}{Le, Q.V.}, \bibinfo{year}{2019}.
\newblock \bibinfo{title}{Efficientnet: Rethinking model scaling for
  convolutional neural networks}.
\newblock \bibinfo{journal}{arXiv preprint arXiv:1905.11946} .
%Type = Misc
\bibitem[{USDA(2019)}]{usda2019}
\bibinfo{author}{USDA}, \bibinfo{year}{2019}.
\newblock \bibinfo{title}{{USDA} long-term agricultural projections}.
\newblock
  \bibinfo{howpublished}{\url{https://www.usda.gov/oce/commodity/projections/}}.
%Type = Article
\bibitem[{Valloli and Mehta(2019)}]{valloli2019w}
\bibinfo{author}{Valloli, V.K.}, \bibinfo{author}{Mehta, K.},
  \bibinfo{year}{2019}.
\newblock \bibinfo{title}{W-net: Reinforced u-net for density map estimation}.
\newblock \bibinfo{journal}{arXiv preprint arXiv:1903.11249} .
%Type = Inproceedings
\bibitem[{Wang et~al.(2015)Wang, Zhang, Yang, Liu and Cao}]{wang2015deep}
\bibinfo{author}{Wang, C.}, \bibinfo{author}{Zhang, H.}, \bibinfo{author}{Yang,
  L.}, \bibinfo{author}{Liu, S.}, \bibinfo{author}{Cao, X.},
  \bibinfo{year}{2015}.
\newblock \bibinfo{title}{Deep people counting in extremely dense crowds}, in:
  \bibinfo{booktitle}{Proceedings of the 23rd ACM international conference on
  Multimedia}, pp. \bibinfo{pages}{1299--1302}.
%Type = Article
\bibitem[{Wang et~al.(2017)Wang, Sun and Wang}]{wang2017automatic}
\bibinfo{author}{Wang, G.}, \bibinfo{author}{Sun, Y.}, \bibinfo{author}{Wang,
  J.}, \bibinfo{year}{2017}.
\newblock \bibinfo{title}{Automatic image-based plant disease severity
  estimation using deep learning}.
\newblock \bibinfo{journal}{Computational intelligence and neuroscience}
  \bibinfo{volume}{2017}.
%Type = Article
\bibitem[{Wang et~al.(2019)Wang, Yin, Tang and Li}]{wang2019removing}
\bibinfo{author}{Wang, L.}, \bibinfo{author}{Yin, B.}, \bibinfo{author}{Tang,
  X.}, \bibinfo{author}{Li, Y.}, \bibinfo{year}{2019}.
\newblock \bibinfo{title}{Removing background interference for crowd counting
  via de-background detail convolutional network}.
\newblock \bibinfo{journal}{Neurocomputing} \bibinfo{volume}{332},
  \bibinfo{pages}{360--371}.
%Type = Inproceedings
\bibitem[{Wu et~al.(2019a)Wu, Zheng, Ye, Hu, Yang and He}]{wu2019adaptive}
\bibinfo{author}{Wu, X.}, \bibinfo{author}{Zheng, Y.}, \bibinfo{author}{Ye,
  H.}, \bibinfo{author}{Hu, W.}, \bibinfo{author}{Yang, J.},
  \bibinfo{author}{He, L.}, \bibinfo{year}{2019}a.
\newblock \bibinfo{title}{Adaptive scenario discovery for crowd counting}, in:
  \bibinfo{booktitle}{ICASSP 2019-2019 IEEE International Conference on
  Acoustics, Speech and Signal Processing (ICASSP)},
  \bibinfo{organization}{IEEE}. pp. \bibinfo{pages}{2382--2386}.
%Type = Article
\bibitem[{Wu et~al.(2019b)Wu, Lin, Dong, Yan, Bian and
  Yang}]{wu2019progressive}
\bibinfo{author}{Wu, Y.}, \bibinfo{author}{Lin, Y.}, \bibinfo{author}{Dong,
  X.}, \bibinfo{author}{Yan, Y.}, \bibinfo{author}{Bian, W.},
  \bibinfo{author}{Yang, Y.}, \bibinfo{year}{2019}b.
\newblock \bibinfo{title}{Progressive learning for person re-identification
  with one example}.
\newblock \bibinfo{journal}{IEEE Transactions on Image Processing}
  \bibinfo{volume}{28}, \bibinfo{pages}{2872--2881}.
%Type = Inproceedings
\bibitem[{Xie et~al.(2020)Xie, Luong, Hovy and Le}]{xie2020self}
\bibinfo{author}{Xie, Q.}, \bibinfo{author}{Luong, M.T.},
  \bibinfo{author}{Hovy, E.}, \bibinfo{author}{Le, Q.V.}, \bibinfo{year}{2020}.
\newblock \bibinfo{title}{Self-training with noisy student improves imagenet
  classification}, in: \bibinfo{booktitle}{Proceedings of the IEEE/CVF
  Conference on Computer Vision and Pattern Recognition}, pp.
  \bibinfo{pages}{10687--10698}.
%Type = Article
\bibitem[{Yuan et~al.(2018)Yuan, Li, Bhatta, Shi, Baenziger and
  Ge}]{yuan2018wheat}
\bibinfo{author}{Yuan, W.}, \bibinfo{author}{Li, J.}, \bibinfo{author}{Bhatta,
  M.}, \bibinfo{author}{Shi, Y.}, \bibinfo{author}{Baenziger, P.S.},
  \bibinfo{author}{Ge, Y.}, \bibinfo{year}{2018}.
\newblock \bibinfo{title}{Wheat height estimation using lidar in comparison to
  ultrasonic sensor and uas}.
\newblock \bibinfo{journal}{Sensors} \bibinfo{volume}{18},
  \bibinfo{pages}{3731}.
%Type = Inproceedings
\bibitem[{Zeng et~al.(2017)Zeng, Xu, Cai, Qiu and Zhang}]{zeng2017multi}
\bibinfo{author}{Zeng, L.}, \bibinfo{author}{Xu, X.}, \bibinfo{author}{Cai,
  B.}, \bibinfo{author}{Qiu, S.}, \bibinfo{author}{Zhang, T.},
  \bibinfo{year}{2017}.
\newblock \bibinfo{title}{Multi-scale convolutional neural networks for crowd
  counting}, in: \bibinfo{booktitle}{2017 IEEE International Conference on
  Image Processing (ICIP)}, \bibinfo{organization}{IEEE}. pp.
  \bibinfo{pages}{465--469}.
%Type = Inproceedings
\bibitem[{Zhang et~al.(2018)Zhang, Shi and Chen}]{zhang2018crowd}
\bibinfo{author}{Zhang, L.}, \bibinfo{author}{Shi, M.}, \bibinfo{author}{Chen,
  Q.}, \bibinfo{year}{2018}.
\newblock \bibinfo{title}{Crowd counting via scale-adaptive convolutional
  neural network}, in: \bibinfo{booktitle}{2018 IEEE Winter Conference on
  Applications of Computer Vision (WACV)}, \bibinfo{organization}{IEEE}. pp.
  \bibinfo{pages}{1113--1121}.
%Type = Inproceedings
\bibitem[{Zhang et~al.(2016)Zhang, Zhou, Chen, Gao and Ma}]{zhang2016single}
\bibinfo{author}{Zhang, Y.}, \bibinfo{author}{Zhou, D.}, \bibinfo{author}{Chen,
  S.}, \bibinfo{author}{Gao, S.}, \bibinfo{author}{Ma, Y.},
  \bibinfo{year}{2016}.
\newblock \bibinfo{title}{Single-image crowd counting via multi-column
  convolutional neural network}, in: \bibinfo{booktitle}{Proceedings of the
  IEEE conference on computer vision and pattern recognition}, pp.
  \bibinfo{pages}{589--597}.
%Type = Article
\bibitem[{Zheng et~al.(2019)Zheng, Kong, Jin, Wang, Su and
  Zuo}]{zheng2019cropdeep}
\bibinfo{author}{Zheng, Y.Y.}, \bibinfo{author}{Kong, J.L.},
  \bibinfo{author}{Jin, X.B.}, \bibinfo{author}{Wang, X.Y.},
  \bibinfo{author}{Su, T.L.}, \bibinfo{author}{Zuo, M.}, \bibinfo{year}{2019}.
\newblock \bibinfo{title}{Cropdeep: the crop vision dataset for
  deep-learning-based classification and detection in precision agriculture}.
\newblock \bibinfo{journal}{Sensors} \bibinfo{volume}{19},
  \bibinfo{pages}{1058}.

\end{thebibliography}

\end{document}